\documentclass[10pt,twocolumn,letterpaper]{article}
\usepackage{cvpr}  
\usepackage[pagebackref,breaklinks,colorlinks]{hyperref}
\usepackage{xcolor}
          
\title{SARMAE: Masked Autoencoder for SAR Representation Learning}

\author{Danxu Liu$^{1,4*}$, Di Wang$^{2,4*}$, Hebaixu Wang$^{3,4*}$, Haoyang Chen$^{2,4*}$, Wentao Jiang$^{2}$\\ Yilin Cheng$^{4,5}$, Haonan Guo$^{4,6}$, Wei Cui$^{1\dagger}$, Jing Zhang$^{2,4\dagger}$\\
$^{1}$School of Information and Electronics, Beijing Institute of
Technology, Beijing, China\\
$^{2}$School of Computer Science, Wuhan University, Wuhan, China\\ 
$^{3}$School of  Electronic Information, Wuhan University, Wuhan, China\\ 
$^{4}$Zhongguancun Academy, Beijing, China\\
$^{5}$School of Data Science, Fudan University, Shanghai, China\\
$^{6}$State Key Laboratory of Information Engineering in Surveying, Mapping and Remote Sensing,\\
Wuhan University, Wuhan, China\\ 
}

\usepackage{multirow}   
\usepackage{booktabs} 
\usepackage{graphicx}
\usepackage{dblfloatfix}
\begin{document}
\maketitle
\renewcommand\thefootnote{*}\footnotetext{Equal contributions.}%
\renewcommand\thefootnote{$\dagger$}\footnotetext{Corresponding authors.}%
\begin{abstract}

Synthetic Aperture Radar (SAR) imagery plays a critical role in all-weather, day-and-night remote sensing applications. However, existing SAR-oriented deep learning is constrained by data scarcity, while the physically grounded speckle noise in SAR imagery further hampers fine-grained semantic representation learning. To address these challenges, we propose SARMAE, a Noise-Aware Masked Autoencoder for self-supervised SAR representation learning. Specifically, we construct SAR-1M, the first million-scale SAR dataset, with additional paired optical images, to enable large-scale pre-training.
Building upon this, we design Speckle-Aware Representation Enhancement (SARE), which injects SAR-specific speckle noise into Masked Autoencoder to facilitate noise-aware and robust representation learning. Furthermore, we introduce Semantic Anchor Representation Constraint (SARC), which leverages paired optical priors to align SAR features and ensure semantic consistency. Extensive experiments across multiple SAR datasets demonstrate that SARMAE achieves state-of-the-art performance on classification, detection, and segmentation tasks. Code, data, and models available at \href{https://github.com/MiliLab/SARMAE}{SARMAE}.

\end{abstract}    
\section{Introduction}
\label{sec:intro}

Synthetic Aperture Radar (SAR) imagery, characterized by its all-weather, day-and-night imaging capability, serves as a critical source of remote sensing information~\cite{zhu2021deep,gens1996review,earthobsv_overview,chan2008introduction, chen2026any2any, krieger2013mimo}. Owing to this robustness against challenging atmospheric and illumination variations, SAR has been widely applied in various domains such as ocean monitoring, disaster assessment, and urban scene analysis~\cite{lv2023recognition,yamaguchi2012disaster,mcnairn2004application,yun2019research,tomiyasu2005tutorial,neagoe2016advanced, TiMo}.

\begin{figure}[t]
  \centering
   \includegraphics[width=\linewidth]{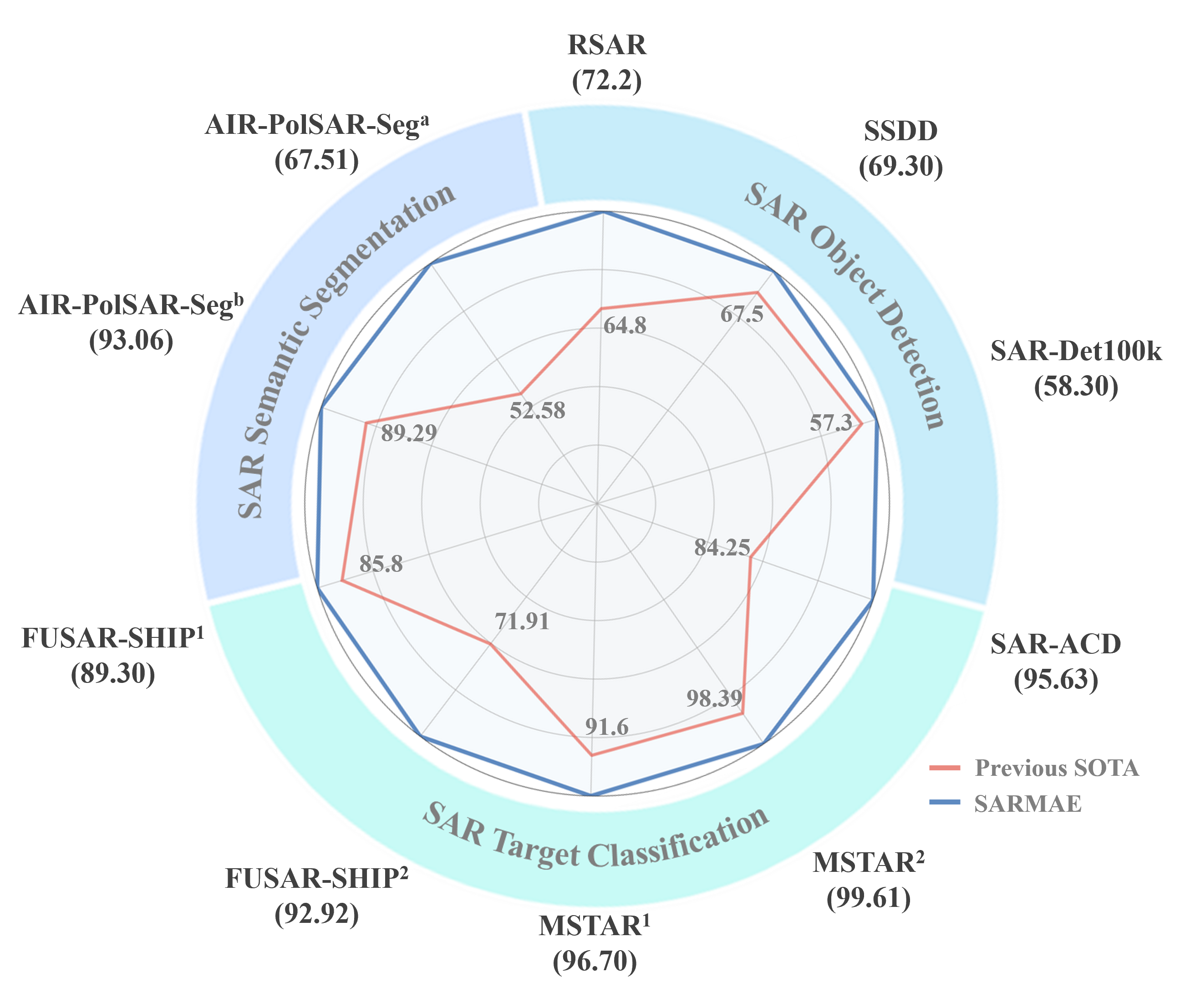}
   \caption{SARMAE outperforms SOTA methods on multiple datasets. $^1$: 40-SHOT; $^2$: 30\% labeled. $^a$: Multi-classes; $^b$: Water.}
   \label{fig:radar}
   \vspace{-5mm}
\end{figure}

The emergence of deep learning has greatly advanced SAR image interpretation~\cite{Zhu2020DeepLM}. Early approaches were typically designed for specific tasks~\cite{era-net}, exhibiting limited generalization across diverse downstream applications. To overcome this limitation, subsequent representation learning methods adopted the pretraining–finetuning paradigm~\cite{rs15235534,DU2025104624,li2024predicting,li2025saratr,10776610} to achieve unified and generalized interpretation. In this context, self-supervised pretraining methods~\cite{mae,beit,simclr} have shown great potential for learning transferable representations from unlabeled data, thereby inspiring the development of analogous frameworks tailored for SAR imagery \cite{saratrx,summit}.

Despite these advances, existing SAR representation learning methods still face several limitations.
First, due to the high cost of SAR data acquisition, current studies often lack sufficiently large and diverse pretraining datasets, limiting the ability of models to learn general-purpose SAR representations~\cite{5206848,mtp,NIPS2012_c399862d,rvsa,Singh2023TheEO,samrs}.
Second, the unique imaging mechanism of SAR introduces severe speckle noise, leading to low semantic content and weak structural cues~\cite{10.1117/12.3037578,krieger2013mimo,article,wang2025residual,wang2025dgsolver,wang2025deep}. However, prevailing pre-training methods simply adopt strategies borrowed from optical imagery~\cite{DU2025104624}, overlooking the physical nature of SAR noise.
Third, pretraining paradigms that rely exclusively on SAR data are inherently limited by its semantic scope~\cite{DBLP:journals/corr/abs-2103-00020,li2023scalinglanguageimagepretrainingmasking,DBLP:journals/corr/abs-2104-02057}. This overlooks the valuable guidance that could be drawn from complementary, information-rich modalities such as optical imagery~\cite{fuller2023cromaremotesensingrepresentations}.
Consequently, the learned representations often lack the semantic richness and generalizability required for diverse downstream tasks.

To address these challenges, we propose SARMAE, a Masked Autoencoder (MAE) \cite{mae}-based framework for SAR representation learning. Specifically, we construct SAR-1M, the first million-scale SAR dataset that covers diverse scenes, targets, and sensor characteristics, providing over 1.3 million SAR images for large-scale pretraining. We then introduce Speckle-Aware Representation Enhancement (SARE), which explicitly incorporates SAR imaging noise during pretraining. By reconstructing the original image from perturbed inputs injected with SAR-specific speckle noise, SARE encourages the network to perceive and capture the intrinsic speckle characteristics of SAR imagery, thereby learning noise-resistant representations and improving robustness. Finally, we design Semantic Anchor Representation Constraint (SARC). Considering the more discernible semantic structure of optical imagery, SARC aligns SAR features with geographically paired optical priors to guide the encoder learning. This alignment enforces semantic consistency, accelerates model convergence, and enhances the generalization of pretrained representations. Extensive experiments on various downstream tasks demonstrate that SARMAE achieves state-of-the-art performance across multiple SAR datasets, as illustrated in Fig.~\ref{fig:radar}. Our main contributions are summarized as follows:

\begin{itemize}
\item[$\bullet$] We propose SAR-1M, the first million-scale SAR dataset covering diverse resolutions, scenes, scales, targets, and task scenarios, along with additional paired optical images, enabling large-scale pretraining.
\item[$\bullet$] We introduce Speckle-Aware Representation Enhancement to enforce robust feature learning by reconstructing inputs corrupted with physically grounded speckle noise, thereby mitigating the adverse impact of noise on SAR feature extraction.
\item[$\bullet$] We design Semantic Anchor Representation Constraint to align SAR features with paired optical priors to regularize encoder learning, ensuring semantic consistency and enhancing representation generalization.
\end{itemize}

\section{Related work}
\label{sec:part2}
\begin{table*}[t]
\centering
\resizebox{\linewidth}{!}{%
\begin{tabular}{lccccccc}
\toprule
\textbf{Model} & \textbf{Pretrain Dataset} & \textbf{Modality} & \textbf{Task Types} & \textbf{Size} &   \textbf{Targets} &  \textbf{Resolution} & \textbf{Band} \\
\midrule
MSFA~\cite{sardet100k} & DOTA~\cite{dota} & OPT & Det. & - & - & - & - \\
SAR-JEPA~\cite{jepa} & SAR-JEPA & SAR & Cls. & 90k & 21 & 0.3m-25m & C/X \\
SARATR-X~\cite{saratrx}& SAR-VSA & SAR & Cls.\&Det. & 180k & 25 & 0.1m-25m & C/X/Ku/Ka \\
SUMMIT~\cite{summit} & MuSID & SAR & Cls.\&Det. & 560k & 41 & 0.1m-25m & C/X \\
SARMAE & SAR-1M & SAR\&OPT & Cls.\&Det.\&Seg. & 1.3M SAR/2.3M Total* & 57 & 0.1m-60m & C/X/Ku/Ka \\
\bottomrule
\end{tabular}
}
\caption{Pretraining datasets for SAR Representation Learning. *Our dataset comprises 1.3M SAR and 1M paired OPT images.}
\label{tab:dataset_comparison}
\end{table*}

\subsection{SAR Datasets for Representation Learning}

Many SAR datasets are characterized by extreme specialization, either focusing on single-category targets such as MSTAR~\cite{mstar} or SSDD~\cite{ssdd}, or being curated for a single downstream task, as exemplified by the detection dataset SAR-Det100k~\cite{sardet100k}.
Beyond this narrow scope lies a more fundamental limitation: the lack of scale across the community’s available resources. Consequently, even state-of-the-art pretrained models such as SARATR-X~\cite{saratrx} and SUMMIT~\cite{summit} are trained on relatively small datasets containing only 180k and 560k images, respectively.
This combination of limited semantic diversity, task specificity, and insufficient data volume imposes a clear ceiling on model performance and hinders the development of generalizable pretraining paradigms for SAR representation learning.
To this end, we introduce SAR-1M, the first million-scale SAR dataset spanning diverse resolutions, imaging scales, object categories, and task scenarios, providing the scale and semantic richness required for large-scale pretraining.

\subsection{SAR Representation Learning} 

Recently, large-scale self-supervised pretraining paradigms have been introduced to remote sensing~\cite{selectivemae, wang2026universal, crossearth, wang2025roma}. SeCo~\cite{seco} adopts a MoCo-style~\cite{moco} framework for scene classification, while RVSA~\cite{rvsa} leverages MAE~\cite{mae} pretraining to obtain initial weights and further introduces rotated varied-size window attention to enhance representation learning. SatMAE~\cite{satmae} extends MAE to multispectral and multitemporal imagery, and ScaleMAE~\cite{scalemae} introduces scale-aware masking to improve generalization across spatial resolutions. Other approaches aim to learn task-agnostic representations by leveraging massive unlabeled datasets across diverse modalities~\cite{ringmo,dofa,croma,wang2025hypersigma,hong2024spectralgpt}. However, these efforts remain predominantly focused on optical imagery, overlooking the unique characteristics of SAR data—particularly its inherent speckle noise and distinct imaging physics.

Motivated by the success of unsupervised pretraining paradigms in optical imagery, several studies have explored self-supervised SAR representation learning. BIDFC~\cite{bidfc} adopts a contrastive learning framework with weak supervision, while SAR-JEPA~\cite{jepa} leverages masked autoencoding and local reconstruction to extract spatial features. MSFA~\cite{sardet100k} further introduces a multi-stage filter-based strategy. Although these approaches provide valuable insights, they remain constrained to single-task settings.
To address these limitations, recent works have proposed unified pretraining frameworks tailored for SAR imagery. SARATR-X~\cite{saratrx} employs a HiViT~\cite{hivit} with a two-stage self-supervised learning framework for unified SAR target detection and classification, while SUMMIT~\cite{summit} adopts a masked image modeling framework with multiple auxiliary self-supervised tasks to adapt to diverse downstream scenarios. These two models demonstrate the feasibility of learning general-purpose SAR representations through large-scale pretraining. However, both methods overlook the intrinsic physical priors of SAR imaging and rely solely on low-quality, noise-contaminated data for pretraining, which limits the semantic richness of their learned representations. In this paper, we propose SARMAE, a noise-aware framework that explicitly models realistic speckle characteristics during pretraining and incorporates high-fidelity paired optical imagery as semantic guidance, thereby enhancing the model’s capability to learn robust and semantically consistent representations from complex SAR data.

\section{Method}
\label{sec:part3}

In this section, we first introduce the constructed SAR-1M dataset.
Based on this foundation, we then present SARMAE, which incorporates two key innovations: Speckle-Aware Representation Enhancement (SARE) and Semantic Anchor Representation Constraint (SARC).

\subsection{SAR-1M Dataset}

Existing public datasets for SAR image analysis are hampered by limitations in scale and task diversity. To overcome these obstacles, we construct a million-scale, multi-source dataset, termed SAR-1M.

\begin{figure}[t]
  \centering
  \includegraphics[width=0.8\linewidth]{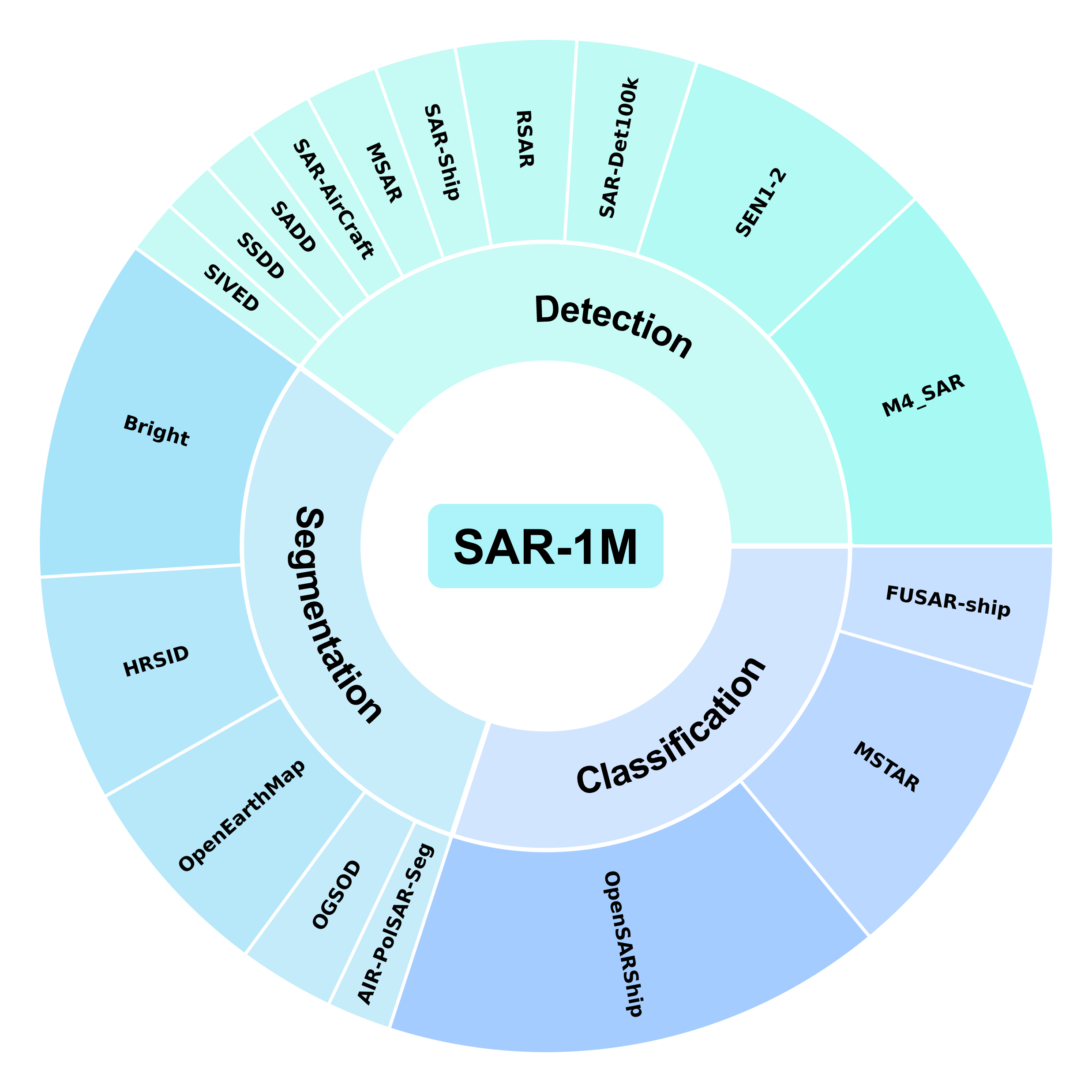}
   \caption{The organization of data sources in SAR-1M.}
   \label{fig:dataset}
   \vspace{-5mm}
\end{figure}

SAR-1M is a curated aggregation of 18 publicly available datasets, as illustrated in Fig.~\ref{fig:dataset}. During data collection, we adhered to two core principles: semantic richness and sensor diversity. For semantic richness, we selected datasets covering representative object categories such as ship, aircraft, vehicle, and bridge, as well as land-cover types including bareland, rangeland, agricultural land, and mountains. In addition, SAR-1M incorporates event-related scenes (\eg, earthquake), further enriching the contextual variety. Notably, many of these datasets were originally designed for detection or segmentation tasks, meaning that their samples typically have larger spatial coverage and richer fine-grained semantics. As a result, SAR-1M encompasses over 50 distinct categories, offering extensive semantic diversity and enhancing the representation capability of SAR-oriented models.

Beyond semantic diversity, SAR-1M integrates imagery from multiple sensors: Sentinel-1, Gaofen-3, RadarSat-2, Terma A/S, and TerraSAR-X, covering multiple frequency bands (C, X, Ku, Ka) and polarization modes (HH, HV, VV, VH). These diverse configurations yield data with varying ground resolutions and scattering characteristics, enriching the representation of different imaging conditions. Moreover, as several source datasets were collected on a global scale (\eg, SEN1-2~\cite{sen12}), SAR-1M exhibits remarkable geographical diversity, which further enhances the generalization and adaptability of pretrained models.

Unlike existing SAR-only datasets, SAR-1M also incorporates geographically aligned SAR–optical pairs, resulting in approximately 2.3 million samples.
Even when restricted to the SAR modality, it still comprises over 1.3 million images, providing ample scale for robust pretraining.

It is worth noting that the number of channels in the originally collected SAR images varies across datasets.
To ensure consistency for unified pretraining, each polarization mode was expanded to a three-channel format when the original image contained a number of channels not equal to three.
A detailed comparison between SAR-1M and existing SAR pretraining datasets is presented in Table~\ref{tab:dataset_comparison}.

\subsection{SARMAE}

\subsubsection{Overall framework}

\begin{figure*}[t]
  \centering
  \includegraphics[width=1.0\linewidth]{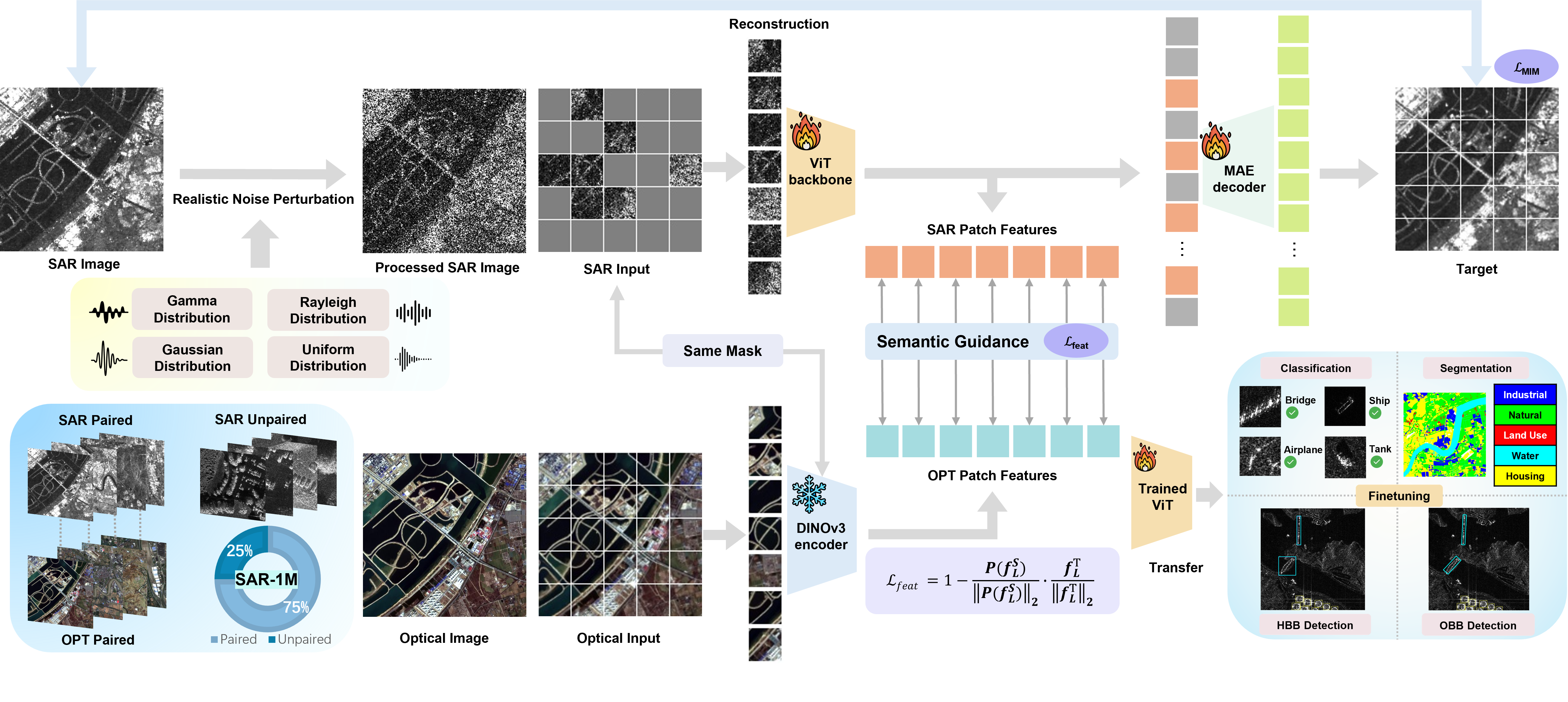}
   \caption{Overview of the SARMAE pretraining framework. The framework consists of two branches: (i) a SAR branch following the MAE architecture with Speckle-Aware Representation Enhancement (SARE) to handle inherent speckle noise, and (ii) an optical branch using a frozen DINOv3 encoder. For paired SAR-optical data, Semantic Anchor Representation Constraint (SARC) aligns SAR features with semantic-rich optical representations. Unpaired SAR images are processed solely through the SAR branch.}
   \label{fig:model}
\end{figure*}

Fig.~\ref{fig:model} illustrates the overall framework of SARMAE for pretraining, which consists of a SAR branch and an optical branch. The SAR branch follows the architecture of MAE~\cite{mae}, where the encoder adopts a ViT backbone and the decoder is composed of multiple Transformer layers same as the original MAE.
The optical branch employs a frozen DINOv3~\cite{dinov3} network that shares the same ViT-based encoder architecture as the SAR branch, facilitating subsequent feature alignment between the two modalities.

During pretraining, a SAR image is fed into the SAR branch, following the same procedure as MAE. To cope with the speckle noise inherent in SAR imagery, we explicitly introduce it into the SAR branch and enforce the network to learn effective denoising representations. This mechanism, termed Speckle-Aware Representation Enhancement (SARE), improves the model’s ability to capture robust and discriminative SAR features.
Furthermore, considering that some SAR images are paired with optical counterparts, we propose a Semantic Anchor Representation Constraint (SARC) to fully exploit such multimodal correspondence. In this design, the optical image is passed through the frozen DINOv3 branch to obtain discriminative semantic features, which serve as semantic anchors to guide the SAR encoder’s representation learning. This constraint not only facilitates model convergence but also enhances generalization capability.

It is worth noting that if a SAR image does not have a paired optical counterpart, it is only processed by the SAR branch. The following sections provide detailed explanations of SARE and SARC.

\subsubsection{Speckle Aware Representation Enhancement}

Unlike additive Gaussian noise, speckle is a multiplicative phenomenon arising from the coherent summation of backscattered signals within a resolution cell. For a multi-look SAR intensity image, the observed intensity $Z$ is the result of incoherently averaging $L$ independent single-look measurements $\{I_i\}_{i=1}^L$, where each $I_i$ follows an exponential distribution with mean $\bar{I}$ equal to the true backscattering intensity. This process is formally described as:
\begin{equation}
Z = \frac{1}{L} \sum_{i=1}^{L} I_i, \quad \text{where } I_i \sim \text{Exp}(\bar{I}), \quad E[I_i] = \bar{I}.
\label{eq:multilook_formation}
\end{equation}
Consequently, the multi-look intensity $Z$ follows a Gamma distribution, with its probability density function (PDF) given by:
\begin{equation}
p_Z(z | \bar{I}, L) = \frac{L^L}{\Gamma(L)\bar{I}^L} z^{L-1} \exp\left(-\frac{Lz}{\bar{I}}\right), \quad z \ge 0,
\label{eq:gamma_distribution}
\end{equation}
where $\Gamma(\cdot)$ is the Gamma function, $L$ is the number of looks, and the mean of $Z$ is $E[Z] = \bar{I}$. Equation \ref{eq:gamma_distribution} provides a physically grounded model for speckle, explicitly showing that the observed pixel intensity $z$ is a random variable drawn from a distribution whose shape is determined by the true scene signal $\bar{I}$. This statistical entanglement is precisely what confuses standard pre-training paradigms and motivates our physics-informed design.

Grounded in this statistical model, we design SARE. Instead of merely reconstruct the original (and already noisy) masked patches, we reformulate the self-supervised task so that the model is trained to denoise a synthetically corrupted version of the input back to the original SAR patch. Specifically, for any given input SAR patch $x$, it is treated as the best available estimate of the true signal (\ie, $x \approx \bar{I}$). A more heavily corrupted observation $x'$ is then synthesized by sampling from the Gamma distribution with a reduced synthetic look number $L_{\text{syn}}$:
\begin{equation}
x'(i,j) \sim \text{Gamma}\left(L_{\text{syn}}, \frac{x(i,j)}{L_{\text{syn}}}\right).
\label{eq:noise_injection}
\end{equation}
This construction preserves the pixel-wise mean $E[x'(i,j)] = x(i,j)$ while increasing the variance through the lower look number. Here, $L_{\text{syn}}$ is chosen to be significantly smaller than the effective number of looks in the original patch, ensuring that the synthesized patch $x'$ exhibits a higher noise level. 

The pre-training task is then formulated as a denoising reconstruction objective. Given the synthetically corrupted patch $x'$, we apply random masking with a high masking ratio (75\% following MAE) to obtain a masked version $\tilde{x}'$, retaining only a subset of visible patches indexed by $\mathcal{V}$. The visible patches are fed to the encoder $E_{SAR}$ to produce latent representations, which are then processed by the decoder $D$ along with learnable mask tokens to reconstruct the original patch $x$. The reconstruction loss of SARE is defined as the mean squared error (MSE) between the predicted patches and the corresponding original patches in the masked regions $\mathcal{M}$:
\begin{equation}
\mathcal{L}_{SARE} = \frac{1}{|\mathcal{M}|} \sum_{p \in \mathcal{M}} \| D(E_{SAR}(\tilde{x}'))_p - x_p \|_2^2,
\label{eq:mae_loss}
\end{equation}
where \(p\) indexes the masked patch positions, \(|\mathcal{M}|\) denotes the number of masked patches, \(D(E_{SAR}(\tilde{x}'))_p\) represents the reconstructed patch at position \(p\), and \(x_p\) is the corresponding ground-truth patch from the original input. This formulation trains the model to map noisy inputs to clean and complete representations. By doing so, the encoder learns features that are robust to the statistical fluctuations inherent to SAR imaging and are largely disentangled from speckle noise.

Notably, in our implementation, in addition to the Gamma model, we also incorporate Rayleigh, Gaussian, and Uniform noise, to generate more diverse corrupted samples and further improve model robustness.

\subsubsection{Semantic Anchor Representation Constraint}

While SARE improves robustness to speckle noise, it does not provide the high-level semantic cues that are readily available in paired optical imagery. To this end, we introduce the SARC.

For a given co-registered image pair $(I_{SAR}, I_{OPT})$, the SAR image is masked using the MAE strategy and passed through $E_{SAR}$ to obtain embeddings for the visible patches,
\begin{equation}
f_{SAR} = \{ f^i_{SAR} \}_{i \in \mathcal{V}},
\label{eq:mae_loss}
\end{equation}
where $\mathcal{V}$ denotes the set of visible patch indices. 
Concurrently, the unmasked optical image is processed by the frozen encoder $E_{OPT}$ to produce a full sequence of anchor embeddings,
\begin{equation}
f_{OPT} = \{ f^i_{OPT} \}_{i=1}^{N}.
\label{eq:mae_loss}
\end{equation}

Then, we enforce that the representation of a SAR patch should be semantically aligned with its spatially corresponding optical patch. This is achieved through a patch-wise cosine distance loss. For each visible SAR patch embedding $f_{SAR}^{i}$ at spatial position $i \in \mathcal{V}$, the optical patch embedding $f_{OPT}^{i}$ at the same location serves as the semantic target. Accordingly, the SARC loss is defined as the mean cosine distance over all visible patches:
\begin{equation}
\mathcal{L}_{SARC} = \frac{1}{|\mathcal{V}|} \sum_{i \in \mathcal{V}} \left(1 - \frac{f_{SAR}^{i} \cdot f_{OPT}^{i}}{\|f_{SAR}^{i}\|_{2} \|f_{OPT}^{i}\|_{2}}\right).
\label{eq:sarc_loss}
\end{equation}

This objective directly minimizes the cosine distance between the SAR and optical feature vectors, enforcing strong directional alignment in the embedding space. As a result, the SAR encoder $E_{SAR}$ learns not only SAR-specific structural cues, but also embeds them in a feature space shaped by the discernible semantic structure provided by the frozen optical teacher.

\textbf{Overall Pre-training Objective.} The overall pre-training loss consists of two components: the SARE loss and the SARC loss:
\begin{equation}
\mathcal{L}_{pretrain} = \mathcal{L}_{SARE} + \lambda\, \mathcal{L}_{SARC},
\label{eq:total_loss}
\end{equation}
\noindent where $\lambda$ balances the modality-specific reconstruction objective and the cross-modal semantic alignment. In our implementation, we set $\lambda = 0.1$. Through this dual-objective optimization, the SAR encoder learns representations that are simultaneously robust to speckle noise and enriched with meaningful semantics, enabling superior performance across downstream perception tasks.

\section{Experiments}
\label{sec:part4}

In this section, we conduct comprehensive experiments to evaluate the proposed method across multiple downstream tasks, including SAR target classification, object detection, and semantic segmentation. We also assess the effectiveness of the SAR-1M dataset and the contributions of the proposed SARE and SARC modules through ablation studies. We initialize the SAR-branch encoder with ViT weights pretrained on ImageNet~\cite{5206848} using MAE~\cite{mae}. 
The model is trained for 300 epochs using AdamW (learning rate = 1$\times$10$^{-3}$, weight decay = 0.05) with a batch size of 1024 and a cosine learning-rate schedule. All the experiments are performed on NVIDIA A800 GPU.
All fine-tuning details are provided in the supplementary material.

\subsection{Main Results}

We compare SARMAE with existing state-of-the-art methods.
In addition to general-purpose visual representation models such as ImageNet-pretrained MAE~\cite{mae}, BEiT~\cite{beit}, and Swin Transformer~\cite{swintransformer}, we also include the most recent SAR-specific approaches, including SAR-JEPA~\cite{jepa}, SARATR-X~\cite{saratrx}, MSFA~\cite{sardet100k}, and SUMMIT~\cite{summit}.

The scene classification and object detection comparison results are obtained from SAR-JEPA~\cite{jepa}, SARATR-X~\cite{saratrx}, and SUMMIT~\cite{summit}, while the semantic segmentation results are directly sourced from AIR-PolSAR-Seg~\cite{zhirui2025air}.
The downstream performance of ViT-B and ViT-L after pretraining is presented in the corresponding tables.

\begin{table}[t]
\centering
\resizebox{\linewidth}{!}{%
\begin{tabular}{l|cc|cc|c}
\toprule
\multirow{2}{*}{\textbf{Method}} & \multicolumn{2}{c|}{\textbf{FUSAR-SHIP}} & 
\multicolumn{2}{c|}{\textbf{MSTAR}} & \textbf{SAR-ACD} \\
\cmidrule{2-6}
                                 & 40-shot & 30\% & 40-shot & 30\% & 30\% \\
\midrule
ResNet-50~\cite{resnet} & - & 58.41 & - & 89.94 & 59.70 \\
Swin Transformer~\cite{swintransformer}    & - & 60.79 & - & 82.97 & 67.50 \\
Beit~\cite{beit}      & 59.70   & 71.13 & 40.70   & 69.75 & 79.77 \\
LoMaR~\cite{lomar}     & 82.70   & - & 77.00     & - & - \\
SAR-JEPA~\cite{jepa}  & 85.80   & - & 91.60   & - & - \\
SUMMIT~\cite{summit}    & - & 71.91 & - & 98.39 & 84.25 \\
Copernicus FM~\cite{Wang_2025_ICCV}    & 87.61 & - & - & - & 92.63 \\
CROMA~\cite{croma}    & 83.71 & - & - & - & 88.99 \\
\midrule
SARMAE(ViT-B)    & 89.30   & \textbf{92.92} & 96.70   & \textbf{99.61} & 95.06 \\
SARMAE(ViT-L)    & \textbf{90.86}   & 92.80 & \textbf{97.24}   & 98.92 & \textbf{95.63} \\
\bottomrule
\end{tabular}%
}
\caption{Performance comparison (Top1 Accuracy, \%) of different methods on the target classification task.}
\label{tab:cls}
\end{table}

\begin{table}[t]
\centering
\resizebox{\linewidth}{!}{%
\begin{tabular}{lcc|lc}
\toprule
\textbf{Method} & \textbf{SARDet-100k} & \textbf{SSDD} & \textbf{Method} & \textbf{RSAR} \\
\midrule
ImageNet~\cite{5206848}         & 52.30 & 66.40  & RoI-Transformer~\cite{roi} & 35.02 \\
Deformable DETR~\cite{detr}  & 50.00 & 52.60  & Def. DETR~\cite{detr}       & 46.62 \\
Swin Transformer~\cite{swintransformer} & 53.80 & 40.70  & RetinaNet~\cite{retinanet}       & 57.67 \\
ConvNeXt~\cite{convnext}         & 55.10 & - & ARS-DETR~\cite{ars}        & 61.14 \\
CATNet~\cite{catnet}           & - & 64.66 & R3Det~\cite{r3det}           & 63.94 \\
MSFA~\cite{sardet100k}             & 56.40 & - & ReDet~\cite{redet}           & 64.71 \\
SARATR-X~\cite{saratrx}           & 57.30 & 67.50  & O-RCNN~\cite{orcnn}          & 64.82 \\
\midrule
SARMAE(ViT-B)    & 57.90 & 68.10  & SARMAE(ViT-B)   & 66.80 \\
SARMAE(ViT-L)    & \textbf{63.10}  & \textbf{69.30}   & SARMAE(ViT-L)   & \textbf{72.20} \\
\bottomrule
\end{tabular}%
}
\caption{Performance comparison (mAP, \%) of different methods on horizontal and oriented object detection tasks.}
\label{tab:det}
\end{table}

\textbf{Target Classification.}
We first evaluate SARMAE on the target classification task, which does not require an additional decoder and therefore provides a direct assessment of the model’s representation capability. In implementation, the feature map from the last ViT layer is processed by global average pooling and then passed through a linear classifier to produce logits corresponding to the number of classes. 
We conduct experiments on three widely used datasets: FUSAR-SHIP~\cite{fusarship}, MSTAR~\cite{mstar}, and SAR-ACD~\cite{saracd}. We compare the performance of SARMAE with Copernicus FM~\cite{Wang_2025_ICCV} and CROMA~\cite{croma}, which adopts DINO distillation and contrastive learning respectively. Table~\ref{tab:cls} shows superior performance of SARMAE on FUSAR and SAR-ACD. Then, following the protocols in~\cite{jepa, summit}, we adopt the 40-shot and 30\% label settings for the FUSAR-SHIP and MSTAR datasets. As shown in Table~\ref{tab:cls}, SARMAE consistently achieves state-of-the-art performance across all evaluation settings. 
For example, under the 30\% label setting on FUSAR-SHIP, our ViT-B model reaches 92.92\% accuracy, outperforming the previous best by a large margin of 21.01\%. 
These results clearly demonstrate the effectiveness of the our approach.

\textbf{Horizontal \& Oriented Object Detection.} We then evaluate the transferability of the learned representations on the object detection task.
For horizontal object detection, we use the classical SSDD~\cite{ssdd} dataset and the more challenging SARDet-100k~\cite{sardet100k}.
For oriented bounding box detection, we adopt the recently proposed large-scale RSAR dataset~\cite{rsar}.
Following the fine-tuning protocols of~\cite{saratrx}, we integrate our pretrained backbones into Faster R-CNN~\cite{fasterrcnn} and Oriented R-CNN~\cite{orcnn} for horizontal and oriented detection, respectively.
The results are detailed in Table~\ref{tab:det}.

In the horizontal object detection scenario represented by SARDet-100k, our model with a standard ViT-B backbone achieves 57.9 mAP, slightly outperforming SARATR-X (built on HiViT-B) which reports 57.3 mAP.
This result is particularly noteworthy because HiViT~\cite{hivit}, with its hierarchical and pyramidal design, is generally regarded as an improved variant of ViT and typically delivers stronger performance.
Nevertheless, our ViT-based model still attains the best results, underscoring that the performance gain primarily arises from the superior representations learned by SARMAE, rather than architectural advantages.

In the oriented object detection task, SARMAE also secures top performance with 66.8\% mAP, surpassing existing methods by a large margin. Notably, our model exhibits excellent scalability: upgrading the backbone from ViT-B to ViT-L further boosts performance by +5.4 mAP, thereby widening the gap over previous approaches.

\textbf{Semantic Segmentation.}
Beyond image-level and object-level evaluation, we further assess the pretrained model on pixel-level tasks through semantic segmentation. For this purpose, we adopt the well-known AIR-PolSAR-Seg dataset~\cite{zhirui2025air}, which includes two settings: multiclass segmentation and single-class water extraction. Our implementation follows the protocol in~\cite{zhirui2025air}, using UperNet~\cite{upernet} as the segmentation framework. As shown in Table~\ref{tab:segmentation_results}, SARMAE achieves clear and consistent advantages over existing methods. By incorporating optical cues through SARC during pretraining, SARMAE captures clearer semantic structures, such as boundaries and texture structures, yielding superior segmentation results, particularly for the more challenging multiclass setting.

\begin{table}[h]
\centering
\resizebox{1.0\linewidth}{!}{%
\small
\begin{tabular}{lccccccc|c}
\toprule
\multirow{2}{*}{\textbf{Method}} & \multicolumn{6}{c}{\textbf{Multiple classes}} & \multirow{2}{*}{\textbf{\ }} & \multicolumn{1}{c}{\textbf{Water}} \\
\cline{2-8}\cline{9-9}
& \textbf{Industrial Area} & \textbf{Natural Area} & \textbf{Land Use} & \textbf{Water} & \textbf{Housing} & \textbf{Other} & \textbf{mIoU} & \textbf{IoU} \\
\midrule
FCN~\cite{long2015fully} & 37.78 & 71.58 & 1.24 & 72.76 & 67.69 & 39.05 & 48.35 & 85.95 \\
ANN~\cite{zhu2019asymmetric} & 41.23 & 72.92 & 0.97 & 75.95 & 68.40 & 56.01 & 52.58 & 87.32 \\
PSPNet~\cite{zhao2017pyramid} & 33.99 & 72.31 & 0.93 & 76.51 & 68.07 & 57.07 & 51.48 & 87.13 \\
DeepLab V3+~\cite{chen2018encoder} & 40.62 & 70.67 & 0.55 & 72.93 & 69.96 & 34.53 & 48.21 & 87.53 \\
PSANet~\cite{Zhao_2018_ECCV} & 40.70 & 69.46 & 1.33 & 69.46 & 68.75 & 32.68 & 47.14 & 86.18 \\
DANet~\cite{fu2019dual} & 39.56 & 72.00 & 1.00 & 74.95 & 67.79 & 56.28 & 39.56 & 89.29 \\
\midrule
SARMAE(ViT-B) & \textbf{65.87} & 75.65 & 29.20 & 84.01 & 73.23 & \textbf{71.21} & 66.53 & 92.31 \\
SARMAE(ViT-L) & 65.84 & \textbf{78.04} & \textbf{29.47} & \textbf{87.12} & \textbf{75.22} & 69.34 & \textbf{67.51} & \textbf{93.06} \\
\bottomrule
\end{tabular}
}
\caption{Performance comparison of semantic segmentation methods on multiple classes and water classes.}
\label{tab:segmentation_results}
\end{table}

\subsection{Ablation Study}

In this section, we investigate the effectiveness of the SAR-1M dataset and conduct ablation studies on SARMAE. We use a ViT-B model pretrained on ImageNet~\cite{5206848} with MAE~\cite{mae} as our baseline to clearly highlight the contribution of each component in our method.
All our ablation models training on SAR-1M use the same settings, i.e., 300 epochs, ensuring a controlled and fair comparison.
For efficient verification, all ablation comparisons are conducted using the ViT-B backbone, and the effectiveness of each variant is demonstrated through fine-tuning on different downstream tasks.

\begin{table}[t]
\centering
\resizebox{1.0\linewidth}{!}{%
\begin{tabular}{lccc|ccc}
\hline
\textbf{Model} & \textbf{Pretrain Dataset} & \textbf{SARE} & \textbf{SARC} & \textbf{FUSAR} & \textbf{SSDD} & \textbf{AIR-PolSAR-Seg} \\
\hline
MAE (Baseline)     & ImageNet-1K~\cite{5206848} &  &  & 75.40  & 64.00  & 60.28 \\
MAE      & MillionAID~\cite{millionaid} &  &  & 80.16  & 63.60  & 62.29 \\
DINOv3      & LVD-1689M~\cite{dinov3} &  &  & 74.25  & 61.60  & 58.42 \\
MAE*             & SAR-1M (only SAR)     &  &  & 82.22 & 64.20  & 64.36 \\
MAE* (Add noise)   & SAR-1M (only SAR)     & \checkmark &   & 86.80  & 64.40  & 65.15 \\
SARMAE*                      & SAR-1M  (SAR/OPT)    & \checkmark & \checkmark & 89.30  & 68.10  & 66.53 \\
\hline
\end{tabular}%
}
\caption{Ablation of SARE and SARC on multiple SAR datasets. *: Initialized with ImageNet pretrained MAE.}
\label{tab:ablation}
\end{table}

\textbf{Efficacy of SAR-1M.} As shown in Table~\ref{tab:ablation}, we evaluate the impact of in-domain pretraining by applying the standard MAE framework across different datasets and fine-tuning on various SAR downstream tasks. It can be seen that, models pretrained from scratch on SAR-1M consistently outperform those pretrained on ImageNet or MillionAID. For example, on the FUSAR-SHIP classification benchmark, in-domain pretraining on SAR-1M improves accuracy from 75.40\% (ImageNet) to 82.22\%, yielding a substantial +6.82\% absolute gain. Similarly, we observe improvements of +0.2 mAP on detection and +4.08 mIoU on segmentation. These results demonstrate that the unique characteristics of SAR imagery introduce a significant distribution gap from both natural and optical remote-sensing images, making it difficult for models pretrained on those modalities to transfer effectively. This underscores the necessity of constructing a large-scale SAR-specific pretraining dataset. Therefore, SAR-1M is not merely a large collection of images, but a high-quality, domain-aligned resource that facilitates the development of powerful representation models tailored for SAR imagery.

\begin{figure}[t]
  \centering
  \includegraphics[width=0.95\linewidth]{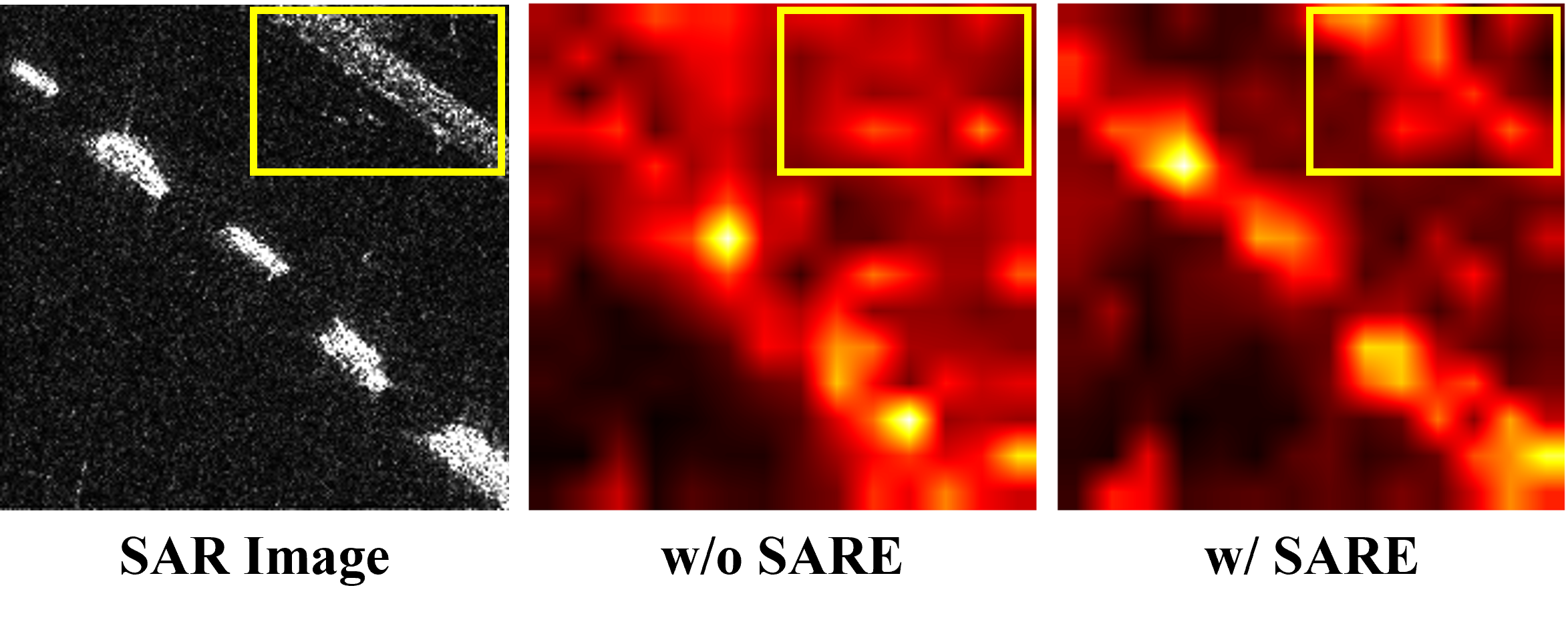}
   \caption{SARE significantly enhances the model’s semantic perception. 
Attention maps are computed by measuring the attention between the MAE encoder’s final-layer class token and image patch tokens.}
   \label{fig:SARE}
\end{figure}

\textbf{Ablation on SARE.} We further validate the effectiveness of the SARE module by integrating it into the SAR-1M–pretrained MAE. As shown in Table~\ref{tab:ablation}, injecting physically grounded speckle noise yields consistent and substantial gains across all downstream tasks: classification accuracy on FUSAR-SHIP improves by +4.58\%, and segmentation mIoU on AIR-PolSARSeg increases by +0.79\%. These results support our central hypothesis that compelling the model to reconstruct denoised targets enables it to better understand the noise characteristics of SAR imagery, thereby achieving noise-aware representation learning. 

To provide more intuitive insights, we qualitatively examine how the model perceives scene content, as shown in Figure~\ref{fig:SARE}. With SARE, scene-relevant objects are more distinctly highlighted and exhibit clearer contours, indicating that the model can overcome noise interference and accurately capture semantic structures. Notably, the model is able to attend to subtle yet semantically relevant objects (see the yellow box), further indicating its enhanced ability to capture meaningful structures in noisy SAR imagery.

\textbf{Ablation on SARC.}
Building upon MAE (add noise), we further evaluate the effectiveness of the SARC module by incorporating paired optical images during pretraining. As shown in Table~\ref{tab:ablation}, SARC yields consistent and significant improvements across all three downstream tasks. The effect is particularly notable on the SSDD detection benchmark, where performance increases by +3.7\% (from 64.4\% to 68.1\%). In our experiments on this dataset, we observe that models often suffer from overfitting manifested as excessive false alarms, likely because the network misinterprets speckle noise as potential targets, indicating insufficient ability to capture true object semantics. By leveraging paired optical images, SARC provides clearer and more discriminative structural cues, enabling the model to better separate meaningful objects from noise. As a result, the detection accuracy improves substantially. Interestingly, although DINOv3 \cite{dinov3} has achieved remarkable performance on natural images, simply finetuning it on SAR images yields suboptimal results. This observation indicates that the effectiveness of SARC stems from explicit SAR–optical alignment, rather than from the strength of the optical backbone itself (i.e., DINOv3).

\begin{figure}[t]
  \centering
  \includegraphics[width=1.0\linewidth]{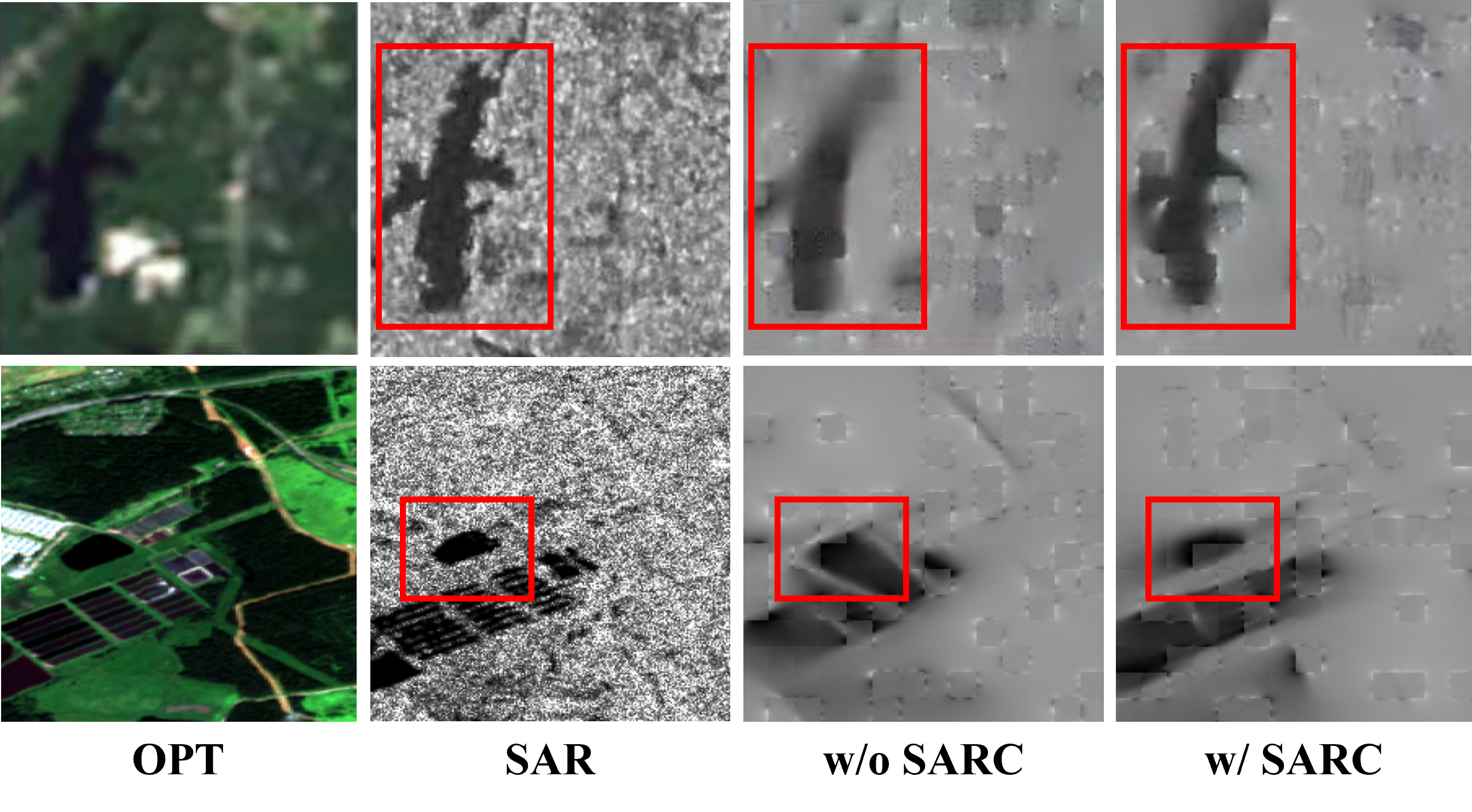}
   \caption{SARC leverages paired optical priors to recover the structural details of the original SAR image, which the model fails to reconstruct without this module.}
   \label{fig:SARC}
\end{figure}

To further illustrate this mechanism, we visualize reconstructions from different pretrained MAE variants, as shown in Figure~\ref{fig:SARC}. With SARC, the model can recover the local scene structures present in the original SAR image (see the red box), whereas training on noisy SAR data alone results in blurred or incomplete reconstructions. This demonstrates that SARC injects clear optical semantic priors, enabling the model to discern true scene content and substantially enhancing the noise-aware masked image modeling process.

Based on the above ablation results, we find that SARE encourages the model to identify noise components irrelevant to scene understanding, enabling it to learn the correct information.
In contrast, SARC introduces paired optical priors that guide the model toward clearer semantic structures.
Together, these two components operate from complementary perspectives and jointly enhance the model’s representation and generalization ability on SAR imagery.

\begin{figure}[t]
  \centering
  \includegraphics[width=1.0\linewidth]{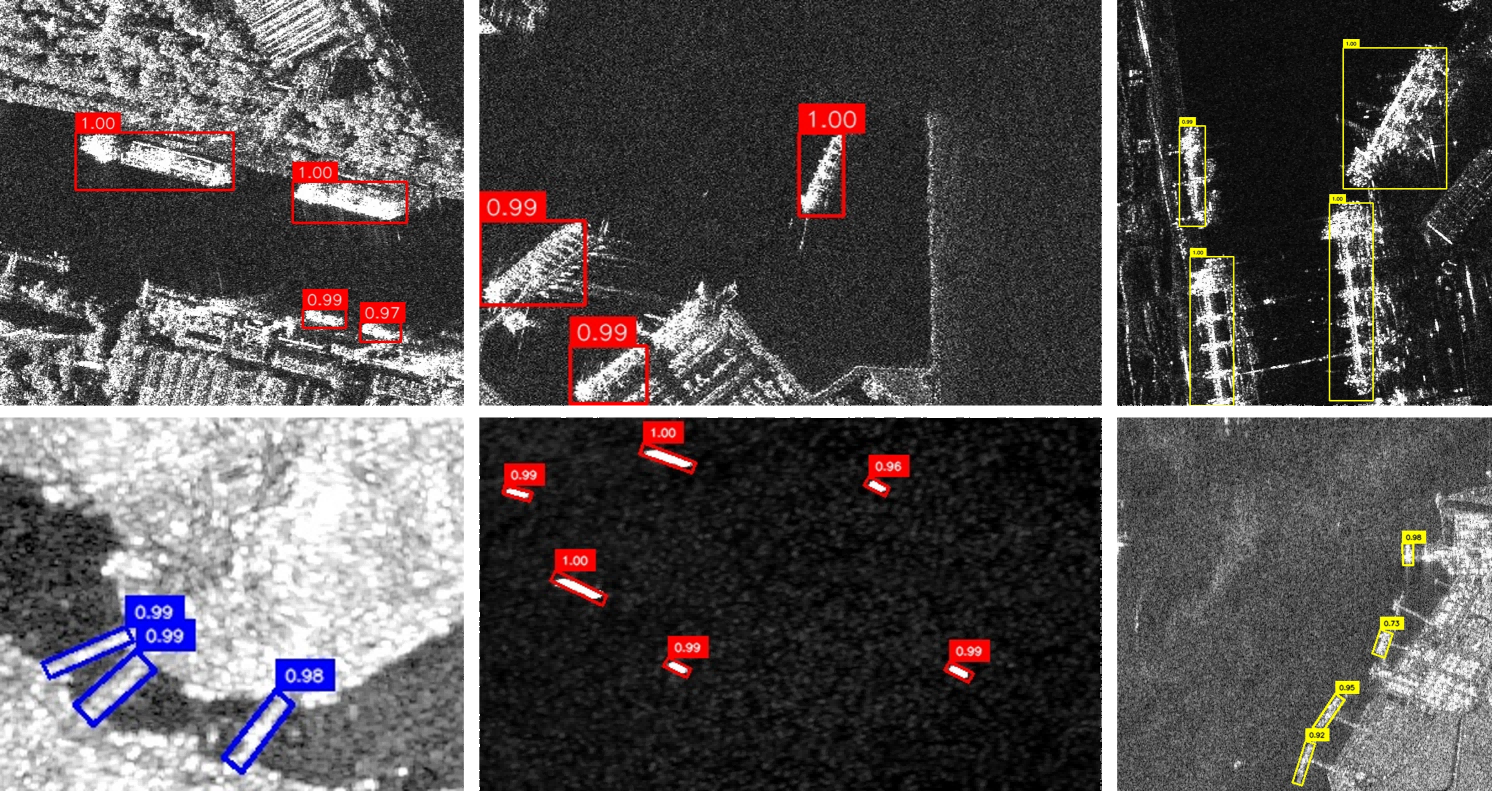}
   \caption{Visualization of Horizontal Object Detection Tasks on SSDD~\cite{ssdd} and SAR-Det100k~\cite{sardet100k} in line 1. Oriented Object Detection Tasks on RSAR~\cite{rsar}.}
   \label{fig:det}
\end{figure}

\begin{figure}[t]
  \centering
  \includegraphics[width=1.0\linewidth]{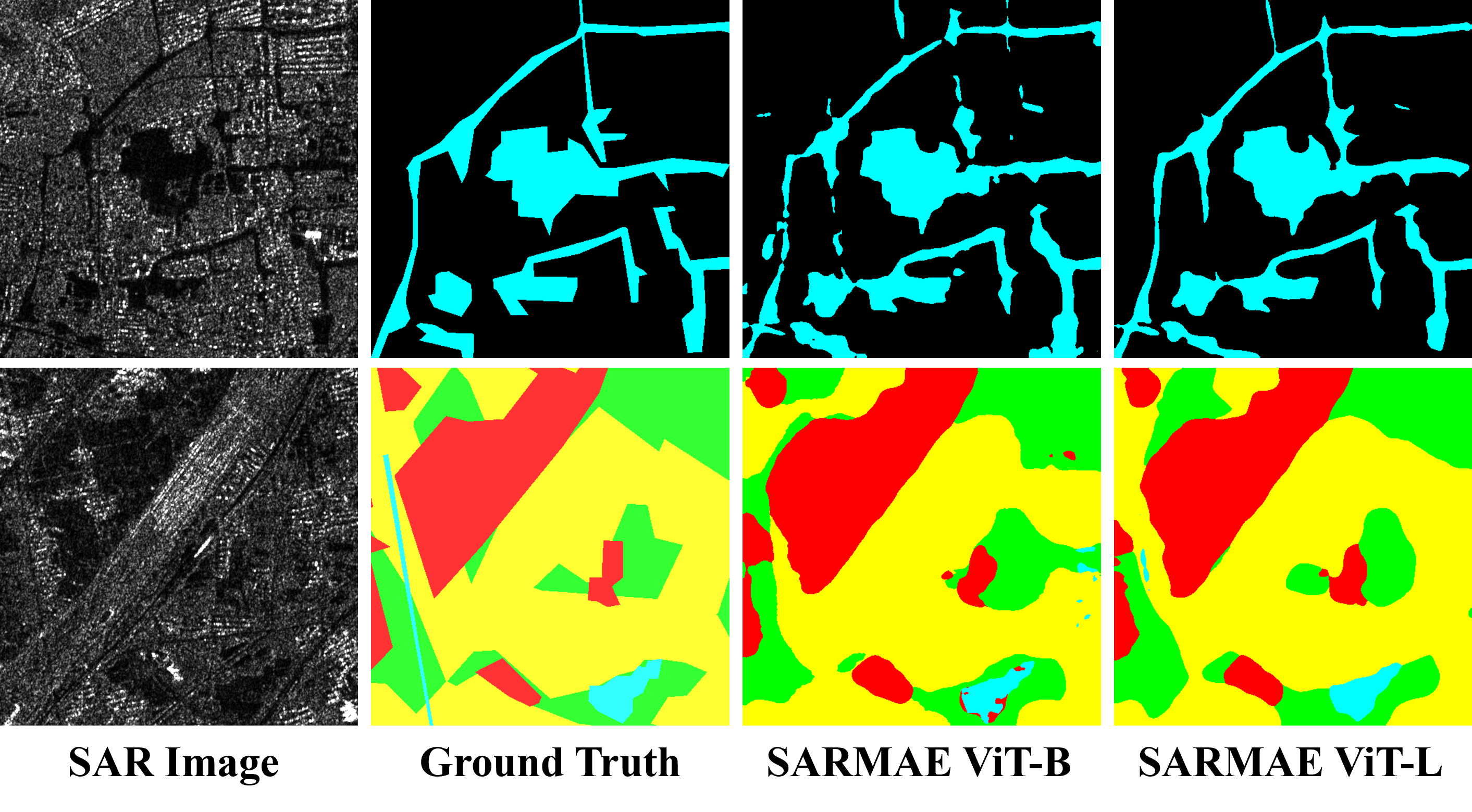}
   \caption{Visualization of segmentation tasks on AIR-PolSAR-Seg~\cite{zhirui2025air}.}
   \label{fig:seg}
\end{figure}

\subsection{Visualization}

Figure~\ref{fig:det}–\ref{fig:seg} present the qualitative results of our model on detection and segmentation tasks, respectively. As shown, our model can accurately detect targets of varying locations, shapes, and sizes in SAR imagery. In particular, for the challenging oriented detection scenario, our approach successfully captures the ship docked at the port with high confidence (see the yellow box in the second row of Figure~\ref{fig:det}), even though its appearance is difficult to distinguish from the surrounding environment. For single-class semantic segmentation, our model achieves robust performance: despite strong noise interference, it is still able to correctly identify water bodies with clear and coherent boundaries. In multi-class segmentation tasks, our method also produces segmentation maps of high visual quality. These results intuitively demonstrate the effectiveness of our approach.
\section{Conclusion}
\label{sec:part5}

In this paper, we present SARMAE, a self-supervised pre-training framework tailored for SAR representation learning. We introduce SAR-1M, the first million-scale SAR dataset with paired optical imagery, enabling large-scale pretraining directly aligned with SAR-specific characteristics. Building upon this, we propose two core innovations: Speckle-Aware Representation Enhancement (SARE), which explicitly models the statistical behavior of SAR speckle to learn noise-aware representations, and Semantic Anchor Representation Constraint (SARC), which leverages paired optical priors to guide SAR feature learning and improve semantic fidelity. Extensive experiments across classification, detection, and segmentation demonstrate that SARMAE achieves SOTA performance, validating both the quality of SAR-1M and the effectiveness of our pre-training framework. We believe that SAR-1M and SARMAE will constitute a strong foundation for future research in SAR representation learning and foster the development of SAR-oriented foundation models.

\section*{Acknowledgement}
This work was supported in part by the Fundamental and Interdisciplinary Disciplines Breakthrough Plan of the Ministry of Education of China (JYB2025XDXM101), the National Natural Science Foundation of China (624B2109, 623B2079, 62225113), the Zhongguancun Academy Project (20240308), and the Key Technology Research Project of China National Petroleum Corporation (2025ZG82).

{
    \small
    \bibliographystyle{ieeenat_fullname}
    \bibliography{main}
}

\clearpage
\setcounter{page}{1}
\maketitlesupplementary

\section{Overview}
\label{sec:rationale}
This supplementary material provides comprehensive details for the proposed SARMAE framework and the constructed SAR-1M dataset. These details were omitted from the main paper due to space constraints.
The supplementary material is organized as follows:
\begin{itemize}
\item Section \ref{sec:sar-1m}. Detailed composition and statistics of SAR-1M dataset;
\item Section \ref{sec:sare}. Implementation details of Speckle-Aware Representation Enhancement (SARE); 
\item Section \ref{sec:finetune}. Fine-tuning configurations for downstream tasks;
\item Section \ref{sec:vis}. Extended visualization results; 
\item Section \ref{app-datasheets}. Datasheet for SAR-1M.
\end{itemize}

\section{Details for SAR-1M.}
\label{sec:sar-1m}

SAR-1M aggregates 18 publicly available SAR datasets, encompassing diverse imaging conditions, geographic locations, and task scenarios. Images in RSAR are the same in SAR-Det100k. Tab.~\ref{tab:dataset} presents the detailed breakdown of each source dataset, including the image quantity, task type, image size, target type and spatial resolution. Datasets with paired SAR\&OPT images are indicated in the last column, comprising 1,042,156 pairs in total.

\begin{table*}[h]
  \caption{Detailed composition of SAR-1M dataset.}
  \label{tab:dataset}
  \centering
  \resizebox{\linewidth}{!}{
  \begin{tabular*}{1\textwidth}{@{\extracolsep{\fill}}lccccccc}
    \toprule
\textbf{Dataset} & \textbf{Year} & \textbf{Tasks} & \textbf{Imgs.} & \textbf{Img. Size (px)} & \textbf{Targets (Cls.)} & \textbf{Res. (m)} & \textbf{Modality} \\
\midrule
    MSTAR\cite{mstar}          & 1995 & Cls.           & 14,577  & 128$\sim$193  & 10             & 0.3      & SAR     \\
    OpenSARShip\cite{opensarship}   & 2017 & Cls.           & 26,679  & 9$\sim$445    & 14            & 2.3$\sim$17.4   & SAR   \\
    SEN1-2\cite{sen12}        & 2018 & Det./Seg.      & 282,384 & 256           & /              & 10        & SAR\&OPT      \\
    SAR-Ship\cite{sarship}       & 2019 & Det.           & 39,729  & 256           & 1              & 3$\sim$25     & SAR   \\
    FUSAR-ship\cite{fusarship}     & 2020 & Cls.           & 5,243   & 512           & 10              & /          & SAR      \\
    HRSID\cite{hrsid}          & 2020 & Det./Seg.      & 89,664  & 256           & 1             & 0.5$\sim$3    & SAR   \\
    SSDD\cite{ssdd}           & 2021 & Det.           & 1,160   & 214$\sim$668  & 1               & 1$\sim$15     & SAR   \\
    SADD\cite{sadd}           & 2022 & Det.           & 2,966   & 224           & 1                & 0.5$\sim$3    & SAR   \\
    MSAR\cite{msar}           & 2022 & Det.           & 28,499  & 256$\sim$2048 & 4                & 1        & SAR        \\
    SAR-AIRcraft\cite{saraircraft}   & 2023 & Det.           & 18,818  & 512           & 7               & 1         & SAR       \\
    SIVED\cite{sived}          & 2023 & Det.           & 1,044   & 512           & 1              & 0.1$\sim$0.3  & SAR   \\
    OGSOD\cite{ogsod}          & 2023 & Det./Seg.      & 22,366  & 256           & 3             & 3          & SAR      \\
    SAR-Det100k\cite{sardet100k}    & 2024 & Det.           & 94,493  & 512           & 6              & 0.5$\sim$25  & SAR    \\
    RSAR\cite{rsar}           & 2025 & Det.           & /       & 512           & 6                & 0.5$\sim$25   & SAR   \\
    M4\_SAR\cite{m4sar}       & 2025 & Det.           & 448,696 & 256           & 6               & 10,60        & SAR\&OPT   \\
    Bright\cite{bright}        & 2025 & Seg.           & 149,872 & 256           & 3               & 0.3$\sim$1   & SAR\&OPT    \\
    OpenEarthMap\cite{xia2025openearthmapsar}   & 2025 & Seg.           & 80,544  & 256           & 8             & 0.15$\sim$0.5  & SAR\&OPT \\
    AIR-PolSAR-Seg\cite{zhirui2025air} & 2025 & Det./Seg.      & 6,168   & 512           & 6               & 8         & SAR       \\
    \midrule
    SAR-1M       & 2025    & Cls./Det./Seg. & 1,312,902 & 9$\sim$2048 & 57                 & 0.1$\sim$60        & SAR\&OPT       \\
    \bottomrule
  \end{tabular*}
  }
\end{table*}

\begin{figure}[t]
  \centering
  \includegraphics[width=0.95\linewidth]{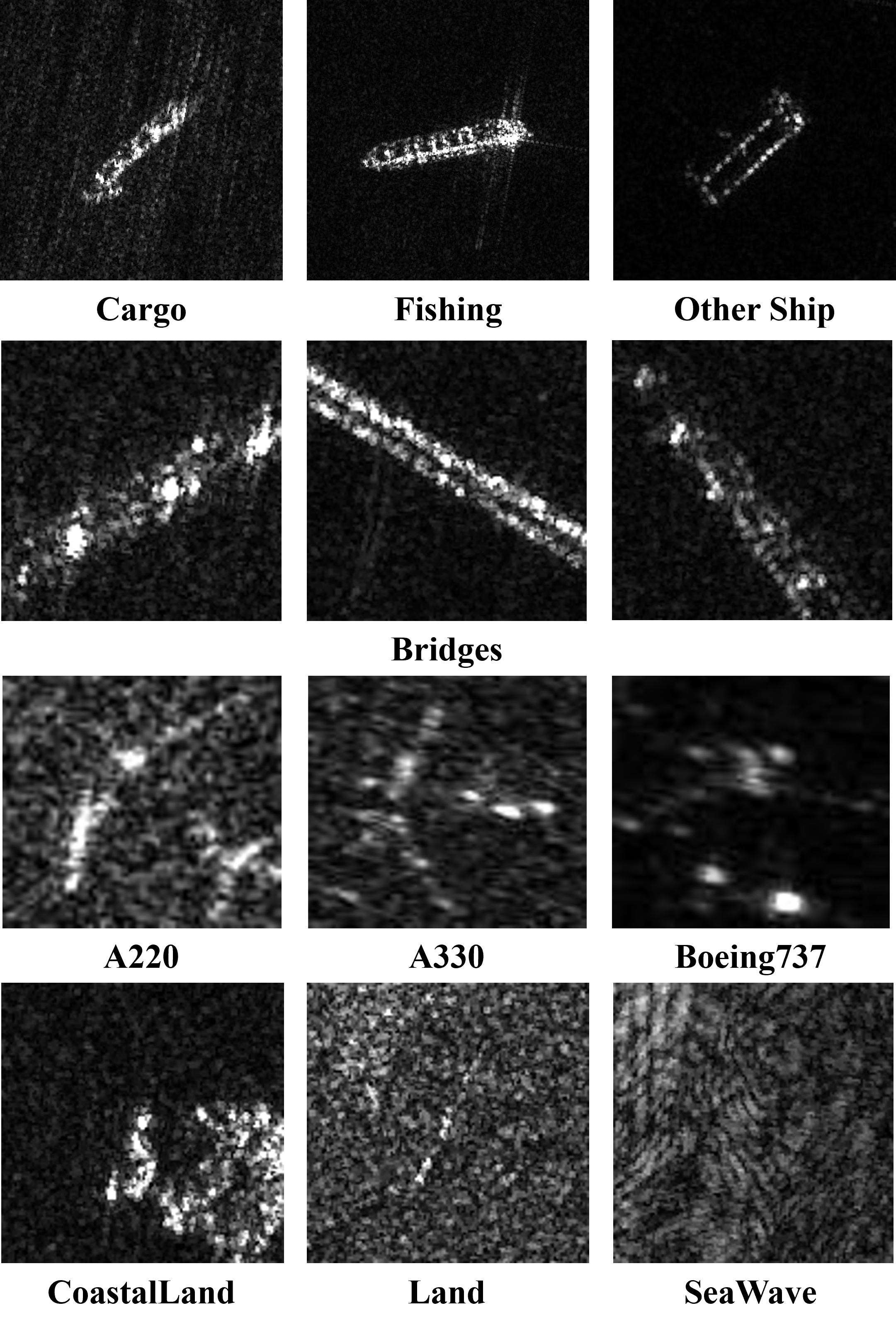}
   \caption{Various categories in SAR-1M.}
   \label{fig:datasetview}
\end{figure}

\begin{figure}[t]
  \centering
  \includegraphics[width=0.95\linewidth]{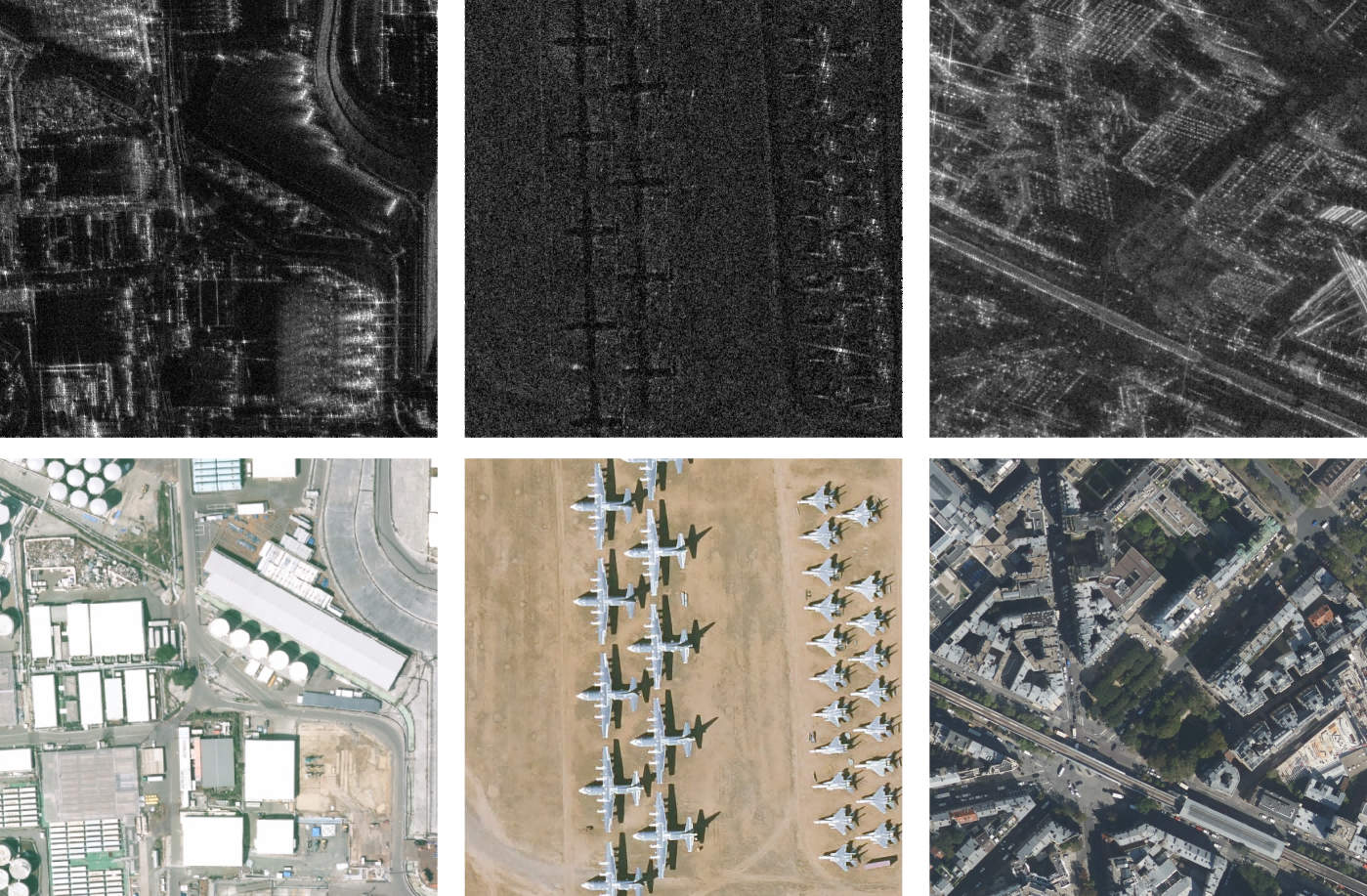}
   \caption{Diverse scenes with SAR-OPT pairs in SAR-1M.}
   \label{fig:optsar}
\end{figure}

SAR-1M encompasses 57 distinct categories, covering a diverse range of scene types. These categories span maritime objects, aerial targets, ground vehicles, infrastructure elements, land cover types, and event-related scenes, with visualizations of different scenarios, either SAR images or SAR–optical paired images, shown in Fig.~\ref{fig:datasetview}-\ref{fig:optsar}. 

\begin{figure*}[t]
  \centering
  \includegraphics[width=0.95\linewidth]{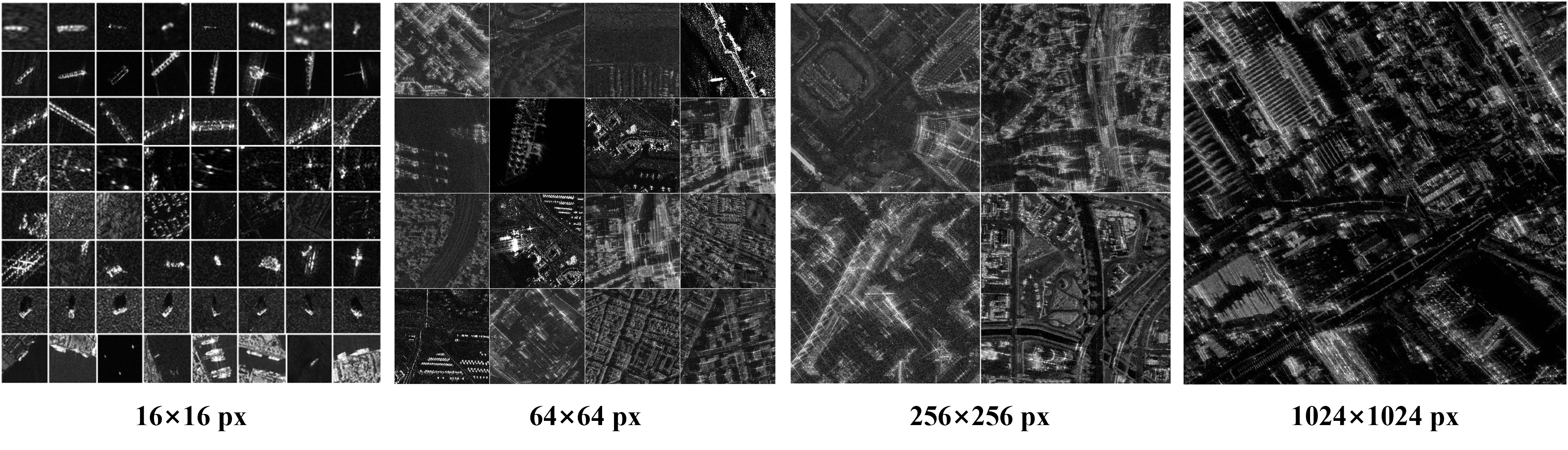}
   \caption{Images of different resolutions in SAR-1M.}
   \label{fig:res}
\end{figure*}

SAR-1M employs hash-based deduplication to eliminate any potential overlap with the test sets of the downstream datasets used in the main paper. Even when trained on only 30\% of SAR-1M (Tab.~\ref{tab:dataset_scale_comparison}), SARMAE-S still outperforms SUMMIT, demonstrating both the stronger capacity of our model and the benefits brought by increased data scale.

\begin{table*}[h]
    \caption{Comparison of performance across different datasets. All models adopt ViT-B as the backbone.}
    \label{tab:dataset_scale_comparison}
    \centering
    \resizebox{0.7\linewidth}{!}{
        \begin{tabular}{lcc|cc}
            \toprule
            \textbf{Model} & \textbf{Dataset} & \textbf{Dataset Scale} & \textbf{FUSAR\_30\%} \cite{fusarship} & \textbf{SAR-ACD} \cite{saracd}\\
            \midrule
            \textcolor{black}{SUMMIT} \cite{summit} & \textcolor{black}{MuSID} & \textcolor{black}{560k} & \textcolor{black}{71.91}  & \textcolor{black}{84.25}\\ 
            SARMAE-S & \textcolor{black}{SAR-1M(30\%)} & \textcolor{black}{300k} & \textcolor{black}{88.27}  & \textcolor{black}{93.76}\\
            \textcolor{black}{SARMAE} & \textcolor{black}{SAR-1M} & \textcolor{black}{1000k} & \textcolor{black}{\textbf{92.92}} & \textcolor{black}{\textbf{95.06}} \\
            \bottomrule
        \end{tabular}
    }
\end{table*}

\section{Implementation Details of SARE}
\label{sec:sare}

To enable the model to learn noise-aware representations while preserving its ability to reconstruct clean SAR imagery, we implement SARE through a carefully designed noise injection strategy during pretraining. Unlike conventional denoising approaches that process all samples uniformly, we adopt a probabilistic augmentation scheme in which each training iteration carries a 50\% chance of applying synthetic noise corruption to the input. This dynamic sampling mechanism allows the encoder to simultaneously learn to reconstruct clean SAR content and to handle diverse noise characteristics.

When an image is selected for augmentation, one of four physically motivated noise models is randomly chosen and applied.

The first category is additive Gaussian noise, which simulates random interference and is defined as:
\begin{equation}
x'(i,j) = x(i,j) + \mathcal{N}(0,\sigma^2),
\end{equation}
where $\sigma$ is randomly sampled from the range $[0.0, 0.5]$ to represent different noise levels.

The second category introduces multiplicative Rayleigh noise, which models the amplitude statistics of single-look SAR data:
\begin{equation}
x'(i,j) = x(i,j) \cdot \mathcal{R}(\sigma),
\end{equation}
where $\mathcal{R}(\sigma)$ denotes a Rayleigh-distributed random variable with scale parameter $\sigma$ sampled from $[0.0, 0.5]$.

The third noise type is Gamma-distributed multiplicative noise, representing the multi-look SAR intensity formation described in Eq.1 of the main text:
\begin{equation}
x'(i,j) \sim \text{Gamma}(L_{\text{syn}}, x(i,j)/L_{\text{syn}}),
\end{equation}
where the synthetic look number $L_{\text{syn}}$ is randomly selected from ${1, 2, 3, 4}$ to yield varying noise intensity levels.

The fourth category is additive uniform noise, which simulates sensor-induced perturbations. It is formulated as:
\begin{equation}
x'(i,j) = x(i,j) + \mathcal{U}(-\alpha, \alpha),
\end{equation}
where $\mathcal{U}(-\alpha, \alpha)$ denotes a uniform random variable drawn from the interval $[-\alpha, \alpha]$, and $\alpha$ is randomly sampled from $[0.0, 0.5]$ to control the perturbation magnitude.

Each chosen noise model further samples its related hyperparameters from the corresponding ranges, ensuring diverse corruption patterns both across iterations and within each training batch.

During training, if augmentation is applied, the corrupted image $x'$ is masked and encoded following the MAE protocol, while the reconstruction target remains the original clean patch $x$. This formulation compels the encoder to map noisy, incomplete inputs back to clean and complete scene content. For iterations where augmentation is skipped, standard MAE reconstruction is performed without synthetic noise. This probabilistic dual-mode training paradigm enables SARMAE to learn noise-robust and semantically rich representations without compromising reconstruction quality or training stability.

Pretraining with pseudo-clean targets generated by SAR denoisers (LEE, BM3D, FROST) yields suboptimal classification accuracy (Tab.~\ref{tab:denoising_mstar}), as such targets often distort semantic structures.
SARE learns to adaptively model complex SAR noise and capture task-relevant representations.
As shown in Tab.~\ref{tab:ablation_study}, SARE outperforms both standard MAE and Gaussian denoising baselines.
These results provide strong evidence that physics-driven noise modeling in SARE offers more robust SAR representations than standard augmentation strategies.

\begin{table}[h]
  \scriptsize
    \caption{Classification accuracy by pretraining on varied denoiser.}
    \label{tab:denoising_mstar}
    \centering
    \resizebox{0.9\linewidth}{!}{
        \begin{tabular}{l|c|ccc}
            \toprule
            \textbf{Denoiser} & \textbf{None} & \textbf{LEE} & \textbf{BM3D} & \textbf{FROST} \\
            \midrule
            MSTAR \cite{mstar} & 90.66 & 77.81 & 82.16 & 76.30 \\
            \bottomrule
        \end{tabular}
    }
\end{table}

\section{Implementation Details of SARC}

We exclude optical images with cloud cover $\ge\!20\%$ to ensure data quality.
To avoid over-regularization from the optical modality, SARC is assigned a low loss weight ($\lambda=0.1$), and additional unpaired SAR samples (0.3M) are incorporated to preserve feature autonomy in the SAR branch.
As shown in Tab.~\ref{tab:ablation_study}, replacing SARC with contrastive loss leads to performance degradation, 
likely because speckle noise in SAR imagery makes negative pairs highly unreliable and interferes with stable feature alignment.
In contrast, cosine similarity focuses on aligning positive pairs without aggressive repulsion, which is more suitable for bridging these distinct modalities.
Moreover, Tab.~\ref{tab:ablation_study} shows that SARE and SARC complement each other, yielding further performance gains when combined.

\begin{table*}[t]
\caption{Ablation study on noise addition and loss functions.}
\label{tab:ablation_study}
\centering
\renewcommand\arraystretch{1.1}
\resizebox{0.85\linewidth}{!}{
\begin{tabular}{lcc|ccc}
\toprule
\textbf{Model} & \textbf{Noise Addition} & \textbf{Loss Function} & \textbf{FUSAR\_40} \cite{fusarship} & \textbf{SSDD} \cite{ssdd} & \textbf{AIR-PolSAR-Seg} \cite{zhirui2025air} \\
\midrule
MAE \cite{mae} & - & - & 82.22 & 64.20 & 64.36 \\
MAE + G & Gaussian & - & 85.16 & 62.60 & 63.19 \\
SARE & Ours & - & 86.80 & \underline{64.40} & \underline{65.15} \\
\midrule
SARE + C & Ours & Contrastive Loss & \underline{86.95} & 62.40 & 63.41 \\
SARMAE & Ours & SARC & \textbf{89.30} & \textbf{68.10} & \textbf{66.53} \\
\bottomrule
\end{tabular}
}
\end{table*}

\section{Rationale for DINOv3}

DINOv3 \cite{dinov3} offers superior optical representations, compared to ImageNet pretrained MAE-ViT (see Tab.~\ref{tab:dino}, both models adopt the Base version).  
We froze DINOv3 encoder to prevent it from adapting to SAR domain, which provides purely semantic guidance in the optical modality.

\begin{table}[h]
\scriptsize
    \caption{Comparison of different teachers for the Optical Branch. ViT-B is adopted as the backbone.}
    \label{tab:dino}
    \centering
    \resizebox{0.9\linewidth}{!}{
        \begin{tabular}{lcc}
            \toprule
            \textbf{Optical Branch Init.} & \textbf{FUSAR\_40} \cite{fusarship}  & \textbf{SAR-ACD} \cite{saracd}\\
            \midrule
            MAE-ViT \cite{mae} & 84.29  &  89.87 \\
            DINOv3 \cite{dinov3} (Ours)         & 89.30   &  95.06 \\
            \bottomrule
        \end{tabular}
    }
\end{table}

\section{Model Scalability Analysis}

In terms of efficiency, scaling from ViT-B to ViT-L introduces a moderate training overhead (+33\%) on 8$\times$A800 GPUs (Tab.~\ref{tab:efficiency}), indicating that SARMAE scales efficiently with model size.
While ViT-L tends to overfit on smaller datasets (FUSAR-SHIP, MSTAR) due to its higher capacity, it consistently outperforms ViT-B on large-scale benchmarks (SARDet-100K).
These results suggest that SARMAE follows standard scaling behavior, where larger models benefit from sufficient training data.

\begin{table}[h]
\scriptsize
    \caption{Efficiency Analysis.}
    \label{tab:efficiency}
    \centering
    \resizebox{\columnwidth}{!}{
        \begin{tabular}{l c c c}
        \toprule  
        \textbf{Model} & \textbf{Params (M)} & \textbf{FLOPs (G)} & \textbf{Training Cost (h)} \\
        \midrule  
        ViT-B (Baseline) & 86  & 17.6 & $\sim$45 \\
        ViT-L            & 307 & 61.6 & $\sim$60 \\
        \bottomrule 
    \end{tabular}
    }
\end{table}

\section{Fine-tuning Configurations for Downstream Tasks}
\label{sec:finetune}

All experiments are conducted using the pretrained ViT-B and ViT-L backbones initialized with SARMAE weights. And all experiments are conducted on 8 NVIDIA A800 GPUs (40GB). We use PyTorch Distributed Data Parallel (DDP) for multi-GPU training. Gradient clipping with a maximum norm of 1.0 is applied across all tasks. For ViT-L models, we apply checkpoints to maintain the effective batch size when GPU memory is limited.

\begin{table*}[htbp]
\centering
\caption{Training configurations for different tasks.}
\label{tab:training_config}
\begin{tabular*}{0.85\textwidth}{@{\extracolsep{\fill}}lccc}
\toprule
\textbf{Config} & \textbf{Classification} & \textbf{Detection} & \textbf{Segmentation} \\
\midrule
optimizer & AdamW & AdamW & AdamW \\
base learning rate & $1.0 \times 10^{-3}$ & $1.0 \times 10^{-4}$ & $6.0 \times 10^{-5}$ \\
weight decay & 0.05 & 0.05 & 0.05 \\
optimizer momentum & $\beta_1, \beta_2 = 0.9, 0.95$ & $\beta_1, \beta_2 = 0.9, 0.95$ & $\beta_1, \beta_2 = 0.9, 0.99$ \\
batch size & 25 & 16 & 4 \\
learning rate schedule & Cosine & Step & Polynomial \\
warmup iterations & 2000 & 1000 & 1500 \\
warmup type & Constant & Linear & Linear \\
warmup learning rate & $1.0 \times 10^{-5}$ & 0.33333 & $6.0 \times 10^{-8}$ \\
\bottomrule
\end{tabular*}
\end{table*}

\subsection{Target Classification}

For target classification tasks, we evaluate SARMAE on three datasets: FUSAR-SHIP~\cite{fusarship}, MSTAR~\cite{mstar}, and SAR-ACD~\cite{saracd}. The pretrained ViT encoder is adapted for classification by appending a global average pooling layer followed by a linear classification head. The number of output dimension in the linear layer corresponds to the number of classes in each dataset. The training configurations are detailed in Tab.~\ref{tab:training_config}. For the 40-shot experiments on FUSAR-SHIP and MSTAR, we randomly sample 40 images per class for training while using the full test set for evaluation. For the 30\% labeled setting on FUSAR-SHIP, MSTAR and SAR-ACD, we randomly select 30\% of the training data while keeping the test set unchanged. Each experiment is repeated 3 times with different random seeds, and we report the average accuracy.

\subsection{Horizontal\&Oriented Object Detection}

For horizontal bounding box detection, we integrate the pretrained SARMAE backbone into the Faster R-CNN \cite{fasterrcnn} framework with a Feature Pyramid Network (FPN) neck. We evaluate on two datasets: SSDD and SARDet-100k. And for oriented bounding box detection on the RSAR dataset, we adopt Oriented R-CNN as the detection framework, which extends Faster R-CNN with rotated Region of Interest (RoI) features and oriented bounding box regression. The training configurations are detailed in Tab.~\ref{tab:training_config}. To preserve the pretrained representations, we freeze all layers of the ViT backbone except the final layer during fine-tuning. This approach maintains the general SAR features learned during pretraining while allowing task-specific adaptation through the detection head.

\subsection{Semantic Segmentation}

For pixel-level semantic segmentation on the AIR-PolSAR-Seg dataset, we utilize UperNet as the segmentation framework. For the multi-class segmentation task, we report mean Intersection over Union (mIoU) across all categories. For the single-class water extraction task, we report the IoU for the water class. The training settings have been shown in Tab.~\ref{tab:training_config}.

\section{Extended visualization results}
\label{sec:vis}

\begin{figure*}[htbp]
  \centering
  \includegraphics[width=0.95\linewidth]{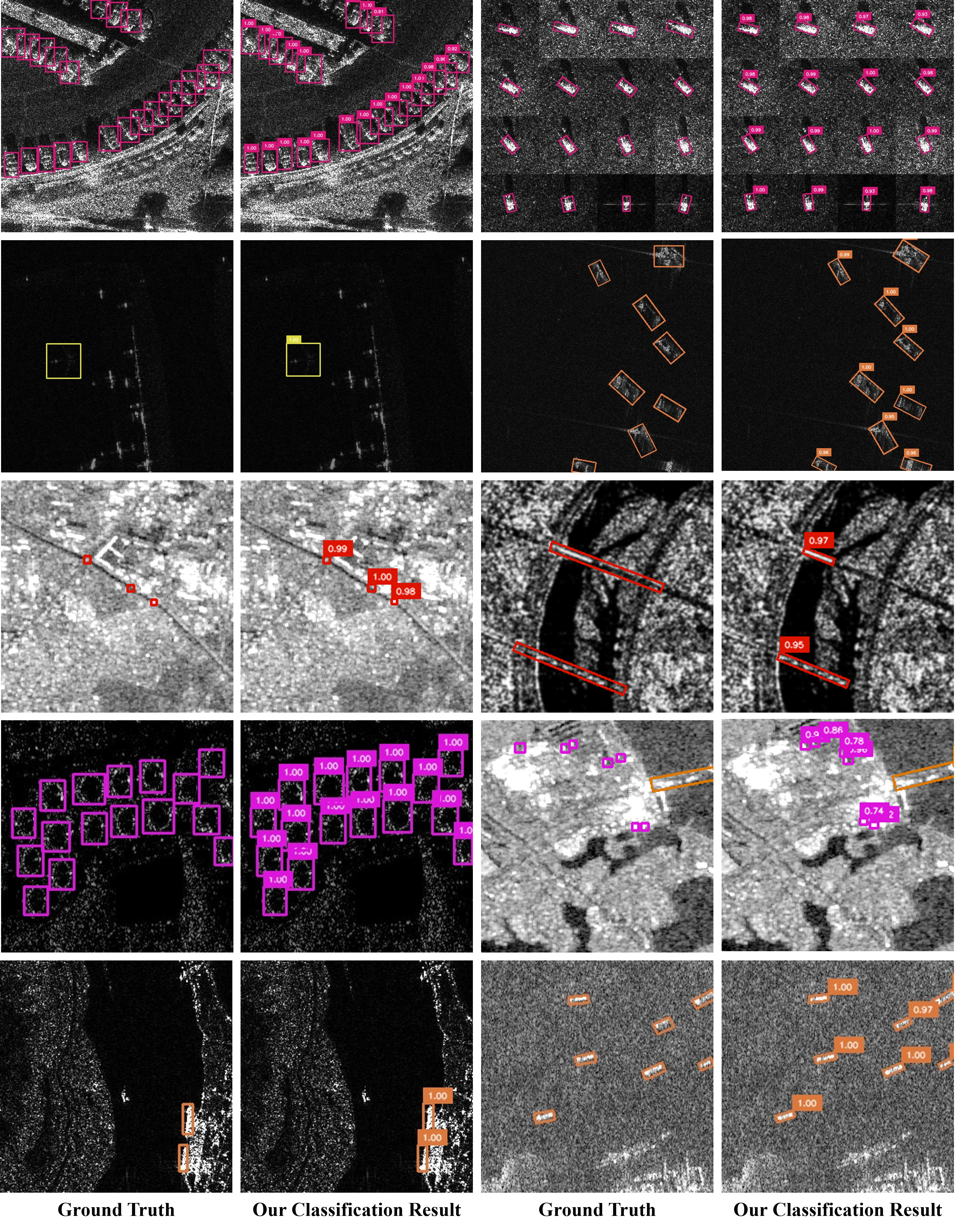}
   \caption{Object detection visualization results.}
   \label{fig:det_vis}
\end{figure*}

Fig.~\ref{fig:det_vis} presents detection results on SSDD, SARDet-100k, and RSAR datasets. The visualizations demonstrate the model's capability to accurately localize ships in diverse scenarios, including multi-scale detection, dense harbor scenes, and oriented bounding box prediction for arbitrarily-oriented vessels.

\begin{figure*}[htbp]
  \centering
  \includegraphics[width=0.9\linewidth]{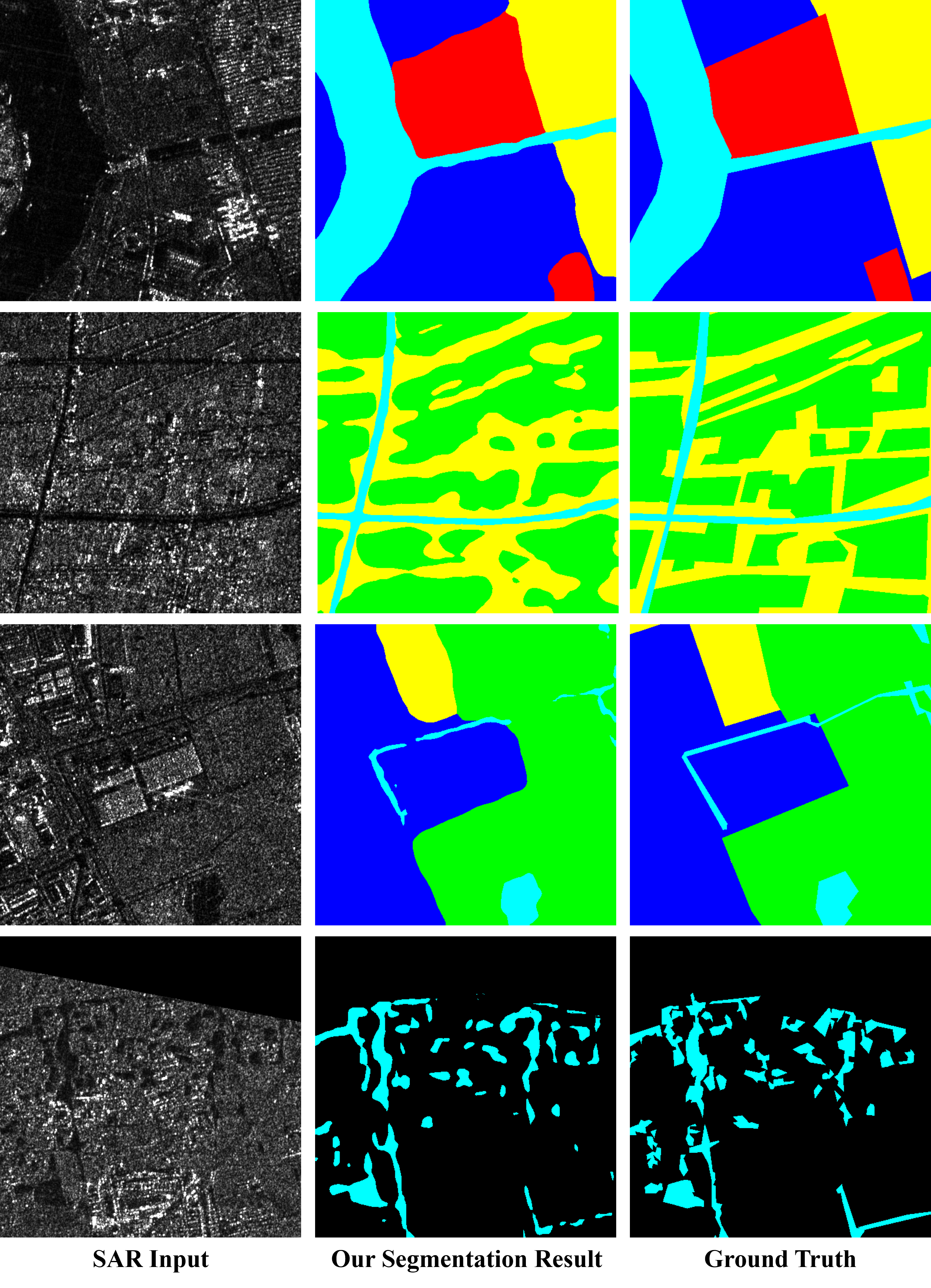}
   \caption{Semantic segmentation visualization results. Blue: Industrial. Green: Natural. Red: Land Use. Cyan: Water. White: Other. Yellow: Housing.}
   \label{fig:seg_vis}
\end{figure*}

Fig.~\ref{fig:seg_vis} illustrates semantic segmentation results on AIR-PolSAR-Seg dataset. The model achieves precise pixel-level classification for multiple terrain categories and accurate water body extraction, demonstrating strong performance on fine-grained segmentation tasks.

\section{Datasheets}
\label{app-datasheets}

\subsection{Motivation}
The questions in this section are primarily intended to encourage dataset creators to clearly articulate their reasons for creating the dataset and to promote transparency about funding interests. The latter may be particularly relevant for datasets created for research purposes.
\begin{enumerate}
    \item \textit{``For what purpose was the dataset created?''}
    
    \textcolor{BurntOrange}{\textbf{A:}} SAR-1M was created to address the lack of large-scale, diverse SAR datasets for self-supervised representation learning. Existing SAR datasets are limited in scale (~100k-500k) and diversity, hindering the development of foundation models for SAR imagery.
    
    \item \textit{``Who created the dataset (\textit{e.g.}, which team, research group) and on behalf of which entity?''}
    
    \textcolor{BurntOrange}{\textbf{A:}} The dataset was curated by us as part of research on SAR representation learning. It aggregates 18 publicly available SAR datasets.

    \item \textit{``Who funded the creation of the dataset?''}
    
    \textcolor{BurntOrange}{\textbf{A:}}
    The dataset creation was funded by the affiliations of the authors involved in this work.
\end{enumerate}

\subsection{Composition}
Most of the questions in this section are intended to provide dataset consumers with the information they need to make informed decisions about using the dataset for their chosen tasks. Some of the questions are designed to elicit information about compliance with the EU’s General Data Protection Regulation (GDPR) or comparable regulations in other jurisdictions. Questions that apply only to datasets that relate to people are grouped together at the end of the section. We recommend taking a broad interpretation of whether a dataset relates to people. For example, any dataset containing text that was written by people relates to people.
\begin{enumerate}
    \item \textit{``What do the instances that comprise our datasets represent (\textit{e.g.}, documents, photos, people, countries)?''}
    
    \textcolor{BurntOrange}{\textbf{A:}} The dataset primarily comprises SAR imagery captured by satellites. All datasets utilized in SAR-1M are publicly accessible and nonprofit.
    
    \item \textit{``How many instances are there in total (of each type, if appropriate)?''}
    
    \textcolor{BurntOrange}{\textbf{A:}} SAR-1M contains 1.3 million SAR image instances captured by satellites and 1 million paired SAR-OPT, 2.3 million in total.

    \item \textit{``Does the dataset contain all possible instances or is it a sample (not necessarily random) of instances from a larger set?''}
    
    \textcolor{BurntOrange}{\textbf{A:}} Yes, our dataset contains all possible instances that have been collected so far.
    
    \item \textit{``Is there a label or target associated with each instance?''}
    
    \textcolor{BurntOrange}{\textbf{A:}} No, our dataset is intended for self-supervised learning. Therefore, each instance is an individual SAR/OPT image and does not contain annotations.
    
    \item \textit{``Is any information missing from individual instances?''}
    
    \textcolor{BurntOrange}{\textbf{A:}} No.
    
    \item \textit{``Are relationships between individual instances made explicit (\textit{e.g.}, users’ movie ratings, social network links)?''}
    
    \textcolor{BurntOrange}{\textbf{A:}} Yes, the relationship between individual instances is explicit.
    
    \item \textit{``Are there recommended data splits (\textit{e.g.}, training, development/validation, testing)?''}
    
    \textcolor{BurntOrange}{\textbf{A:}} Yes, the entire dataset is intended for self-supervised methods, and we recommend using the whole dataset for self-supervised learning research.
    
    \item \textit{``Is the dataset self-contained, or does it link to or otherwise rely on external resources (\textit{e.g.}, websites, tweets, other datasets)?''}
    
    \textcolor{BurntOrange}{\textbf{A:}} Yes, our dataset relies on many publicly available SAR datasets, which we have detailed in the main text.
    
    \item \textit{``Does the dataset contain data that might be considered confidential (\textit{e.g.}, data that is protected by legal privilege or by doctor–patient confidentiality, data that includes the content of individuals’ non-public communications)?''}
    
    \textcolor{BurntOrange}{\textbf{A:}} No, all data are clearly licensed.
    
    \item \textit{``Does the dataset contain data that, if viewed directly, might be offensive, insulting, threatening, or might otherwise cause anxiety?''}
    
    \textcolor{BurntOrange}{\textbf{A:}}
    No.
\end{enumerate}

\subsection{Collection Process}
In addition to the goals outlined in the previous section, the questions in this section are designed to elicit information that may help researchers and practitioners create alternative datasets with similar characteristics. Again, questions that apply only to datasets that relate to people are grouped together at the end of the section.
\begin{enumerate}
    \item \textit{``How was the data associated with each instance acquired?''}
    
    \textcolor{BurntOrange}{\textbf{A:}} Please refer to the details listed in the main text Sec. 2.
    
    \item \textit{``What mechanisms or procedures were used to collect the data (\textit{e.g.}, hardware apparatuses or sensors, manual human curation, software programs, software APIs)?''}
    
    \textcolor{BurntOrange}{\textbf{A:}} Please refer to the details listed in the main text Sec. 2.
    
    \item \textit{``If the dataset is a sample from a larger set, what was the sampling strategy (\textit{e.g.}, deterministic, probabilistic with specific sampling probabilities)?''} 
    
    \textcolor{BurntOrange}{\textbf{A:}} Please refer to the details listed in the main text Sec. 2.
\end{enumerate}

\subsection{Preprocessing, Cleaning, and Labeling}
The questions in this section are intended to provide dataset
consumers with the information they need to determine whether the “raw” data has been processed in ways that are compatible with their chosen tasks. For example, text that has been converted into a ``bag-of-words" is not suitable for tasks involving word order.
\begin{enumerate}
    \item \textit{``Was any preprocessing/cleaning/labeling of the data done (\textit{e.g.}, discretization or bucketing, tokenization, part-of-speech tagging, SIFT feature extraction, removal of instances, processing of missing values)?''}
    
    \textcolor{BurntOrange}{\textbf{A:}} Yes, we preprocessed and cleaned data in our dataset.

    \item \textit{``Was the `raw' data saved in addition to the preprocessed/cleaned/labeled data (\textit{e.g.}, to support unanticipated future uses)?''} 
    
    \textcolor{BurntOrange}{\textbf{A:}} Yes, raw data is accessible.
    
    \item \textit{``Is the software that was used to preprocess/clean/label the data available?''} 
    
    \textcolor{BurntOrange}{\textbf{A:}} Yes, the necessary software used to preprocess and clean the data is publicly available.
\end{enumerate}

\subsection{Uses}
The questions in this section are intended to encourage dataset creators to reflect on tasks for which the dataset should and should not be used. By explicitly highlighting these tasks, dataset creators can help dataset consumers make informed decisions, thereby avoiding potential risks or harms.
\begin{enumerate}
    \item \textit{``Has the dataset been used for any tasks already?''} 
    
    \textcolor{BurntOrange}{\textbf{A:}} No.
    
    \item \textit{``Is there a repository that links to any or all papers or systems that use the dataset?''} 
    
    \textcolor{BurntOrange}{\textbf{A:}} Not yet, but we will provide such links in our GitHub repository soon in the future.
    
    \item \textit{``What (other) tasks could the dataset be used for?''} 
    
    \textcolor{BurntOrange}{\textbf{A:}} The dataset could be used for training the SAR foundation models with the self-supervised learning method.
    
    \item \textit{``Is there anything about the composition of the dataset or the way it was collected and preprocessed/cleaned/labeled that might impact future uses?''} 
    
    \textcolor{BurntOrange}{\textbf{A:}} N/A.
    
    \item \textit{``Are there tasks for which the dataset should not be used?''} 
    
    \textcolor{BurntOrange}{\textbf{A:}} N/A.
\end{enumerate}

\subsection{Distribution}
Dataset creators should provide answers to these questions prior to distributing the dataset either internally within the entity on behalf of which the dataset was created or externally to third parties.
\begin{enumerate}
    \item \textit{``Will the dataset be distributed to third parties outside of the entity (\textit{e.g.}, company, institution, organization) on behalf of which the dataset was created?''} 
    
    \textcolor{BurntOrange}{\textbf{A:}} No.
    
    \item \textit{``How will the dataset be distributed (\textit{e.g.}, tarball on website, API, GitHub)?''} 
    
    \textcolor{BurntOrange}{\textbf{A:}} Very likely to be distributed by website, API, and GitHub repository.
    
    \item \textit{``When will the dataset be distributed?''} 
    
    \textcolor{BurntOrange}{\textbf{A:}} The datasets are publicly accessible, our SAR-1M will be publicly available soon.
    
    \item \textit{``Will the dataset be distributed under a copyright or other intellectual property (IP) license, and/or under applicable terms of use (ToU)?''} 
    
    \textcolor{BurntOrange}{\textbf{A:}} Yes, the dataset is under the Creative Commons Attribution-NonCommercial-ShareAlike 4.0 International License.
    
    \item \textit{``Have any third parties imposed IP-based or other restrictions on the data associated with the instances?''} 
    
    \textcolor{BurntOrange}{\textbf{A:}} No.
    
    \item \textit{``Do any export controls or other regulatory restrictions apply to the dataset or to individual instances?''} 
    
    \textcolor{BurntOrange}{\textbf{A:}} No.    
\end{enumerate}

\subsection{Maintenance}
As with the questions in the previous section, dataset creators should provide answers to these questions prior to distributing the dataset. The questions in this section are intended to encourage dataset creators to plan for dataset maintenance and communicate this plan to dataset consumers.
\begin{enumerate}
    \item \textit{``Who will be supporting/hosting/maintaining the dataset?''} 
    
    \textcolor{BurntOrange}{\textbf{A:}} The authors of this work serve to support, host, and maintain the datasets.
    
    \item \textit{``How can the owner/curator/manager of the dataset be contacted (\textit{e.g.}, email address)?''} 
    
    \textcolor{BurntOrange}{\textbf{A:}} The curators can be contacted via the email addresses listed on our webpage.
    
    \item \textit{``Is there an erratum?''} 
    
    \textcolor{BurntOrange}{\textbf{A:}} There is no explicit erratum; updates and known errors will be specified in future versions.
    
    \item \textit{``Will the dataset be updated (\textit{e.g.}, to correct labeling errors, add new instances, delete instances)?''} 
    
    \textcolor{BurntOrange}{\textbf{A:}} Yes, for the current version. Future updates (if any) will be posted on the dataset website.
    
    \item \textit{``Will older versions of the dataset continue to be supported/hosted/maintained?''} 
    
    \textcolor{BurntOrange}{\textbf{A:}} Yes. This is the first version of the release; future updates will be posted and older versions will be replaced.
    
    \item \textit{``If others want to extend/augment/build on/contribute to the dataset, is there a mechanism for them to do so?''} 
    
    \textcolor{BurntOrange}{\textbf{A:}} Yes, we provide detailed instructions for future extensions.
\end{enumerate}


\end{document}